\documentclass[sigconf,nonacm]{acmart}

\usepackage{amssymb}
\usepackage{multirow}
\usepackage{tikz}
\usepackage{listings}

\lstdefinestyle{pycode}{
  language=Python,
  basicstyle=\ttfamily\scriptsize,
  keywordstyle=\color{blue!55!black}\bfseries,
  commentstyle=\color{green!40!black}\itshape,
  stringstyle=\color{purple!55!black},
  showstringspaces=false,
  numbers=left, numberstyle=\tiny\color{gray!80}, numbersep=7pt,
  breaklines=true, breakatwhitespace=false,
  postbreak=\mbox{\textcolor{gray}{$\hookrightarrow$}\space},
  tabsize=4, columns=fullflexible, keepspaces=true,
  frame=tb, framerule=0.6pt, aboveskip=1em, belowskip=1em,
  backgroundcolor=\color{gray!7},
  captionpos=b,
}

\definecolor{medalgold}{HTML}{E2B007}
\definecolor{medalsilver}{HTML}{9AA0A6}
\definecolor{medalbronze}{HTML}{C77B30}
\DeclareRobustCommand{\medal}[2]{\tikz[baseline=-0.55ex]{%
  \node[circle,fill=#1,inner sep=0pt,minimum size=1.55ex]
       {\textcolor{white}{\fontsize{4.6}{4.6}\selectfont\bfseries #2}};}}
\DeclareRobustCommand{\gmedal}{\medal{medalgold}{1}}
\DeclareRobustCommand{\smedal}{\medal{medalsilver}{2}}
\DeclareRobustCommand{\bmedal}{\medal{medalbronze}{3}}

\begin{document}

\title{Have an LLM Write Your Anomaly Detector: Autonomous Discovery of Compact, Interpretable Detectors for Time Series}

\author{David Berghaus}
\affiliation{\institution{Lamarr Institute}\country{Germany}}
\affiliation{\institution{Fraunhofer IAIS}\country{Germany}}
\email{david.berghaus@iais.fraunhofer.de}
\renewcommand{\shortauthors}{Berghaus}
\settopmatter{printccs=false,printacmref=false}

\begin{abstract}
Time-series anomaly detection trades off predictive accuracy, computational efficiency, and interpretability.
We use a large language model not as the detector but as the \emph{author} of one: an autonomous research loop in which the model repeatedly edits a single short NumPy program under a leakage-free objective, keeping the best-scoring detector it finds.
The loop discovers two compact detectors, one for univariate and one for multivariate series, that describe short windows by their local spectral features and compare them with the training-region distribution through a covariance-aware distance.
On the TSB-AD benchmark these detectors lead the field across metrics, ahead of the strongest classical, deep, and foundation-model baselines including Time-RCD, yet they train no network and use no GPU, and the multivariate detector is faster than every similarly performing baseline.
LLM-driven program search is thus a practical route to accurate, efficient, and transparent detectors.
\end{abstract}




\maketitle

\section{Introduction}

Anomaly detection in time series underpins a long list of practical systems, from monitoring industrial machinery and spacecraft telemetry to flagging irregular heartbeats and intrusions in network traffic.
Despite sustained attention, the field has not converged on a clear winner.
In multivariate settings, deep models can be more competitive, especially when cross-channel structure matters, but strong classical methods such as KMeans and KNN remain among the leading baselines~\citep{qiu2025tab,liuelephant}.
No single method dominates across datasets, anomaly types, and computational budgets.

Classical detectors are cheap and simple but often require dataset-specific preprocessing and threshold tuning to perform well.
Deep and foundation models reduce the manual tuning burden but pay for it with training cost, GPU dependence, and opacity.

We take a different angle: rather than designing a detector by hand or training a large model, we use an \emph{LLM-driven autonomous research loop} as a tool to write the detector for us.
A language model repeatedly edits a single short Python and NumPy program.
It is given the benchmark harness, a strict leakage-free protocol, and an instruction to find an accurate, elegant, single-principle detector.
Each edited program is run through the harness and scored on a held-out validation split; this score, together with the whole trajectory of past attempts (including edits that lowered it), informs the next edit.

Using a language model as an engine for proposing and refining programs, rather than as the final model, underlies a growing line of work in mathematical and algorithmic discovery~\citep{romeraparedes2023mathematical,novikov2025alphaevolve,lange2025shinka,berghaus2026evil}.
Our question is a practical one: is this tool effective for time-series anomaly detection, and what kind of detector does it produce?
As ablations we also hold the harness, the seed program (a simple z-score detector), and the single-principle prompt \emph{fixed} and vary the search by driving the loop with other models (GPT 5.6 Terra and the smaller open MiniMax-2.5) and comparing against an off-the-shelf evolutionary program-search framework, \textsc{ShinkaEvolve}~\citep{lange2025shinka}, which is an open-source alternative to AlphaEvolve~\citep{novikov2025alphaevolve}.

The search does not require existing labels: candidates are scored against synthetic anomalies injected into a slice of the training data, never against real labels, so the approach is also applicable to real-world problems where no ground truth exists.
At inference the discovered detector is fit per series without labels, under the same operating assumptions as a classical unsupervised detector such as KMeans.
We steer the search toward short, efficient programs, and run it separately for univariate and multivariate series.

Our contributions are as follows.
\begin{itemize}
  \item \textbf{An LLM that writes the detector.} We show that an LLM-driven autonomous research loop is an effective tool for discovering anomaly detectors. The search is straightforward to set up, demands little anomaly-detection expertise from the practitioner, and produces a detector that is compact, interpretable, and competitive with the state of the art.
  Each detector emerged from a \textbf{single one hour session} of roughly $100$ edit-and-evaluate iterations.
  \item \textbf{Interpretable, single-CPU programs.} Each detector is a short NumPy program with no GPU, no neural training, and no learned weights, so its inductive biases can be read directly from the source and audited.
  \item \textbf{Field-leading accuracy.} On the established \textbf{TSB-AD} benchmark the two detectors take the most first-place finishes on every reported metric across both regimes, ahead of classical baselines as well as deep and pretrained detectors including the Time-RCD foundation model~\citep{lan2026timercd} (Table~\ref{tab:timercd-protocol-full}), and also perform competitively on the recent TAB benchmark (Appendix~\ref{app:full}).
\end{itemize}

\begin{figure*}[t]
  \centering
  \includegraphics[width=0.85\textwidth]{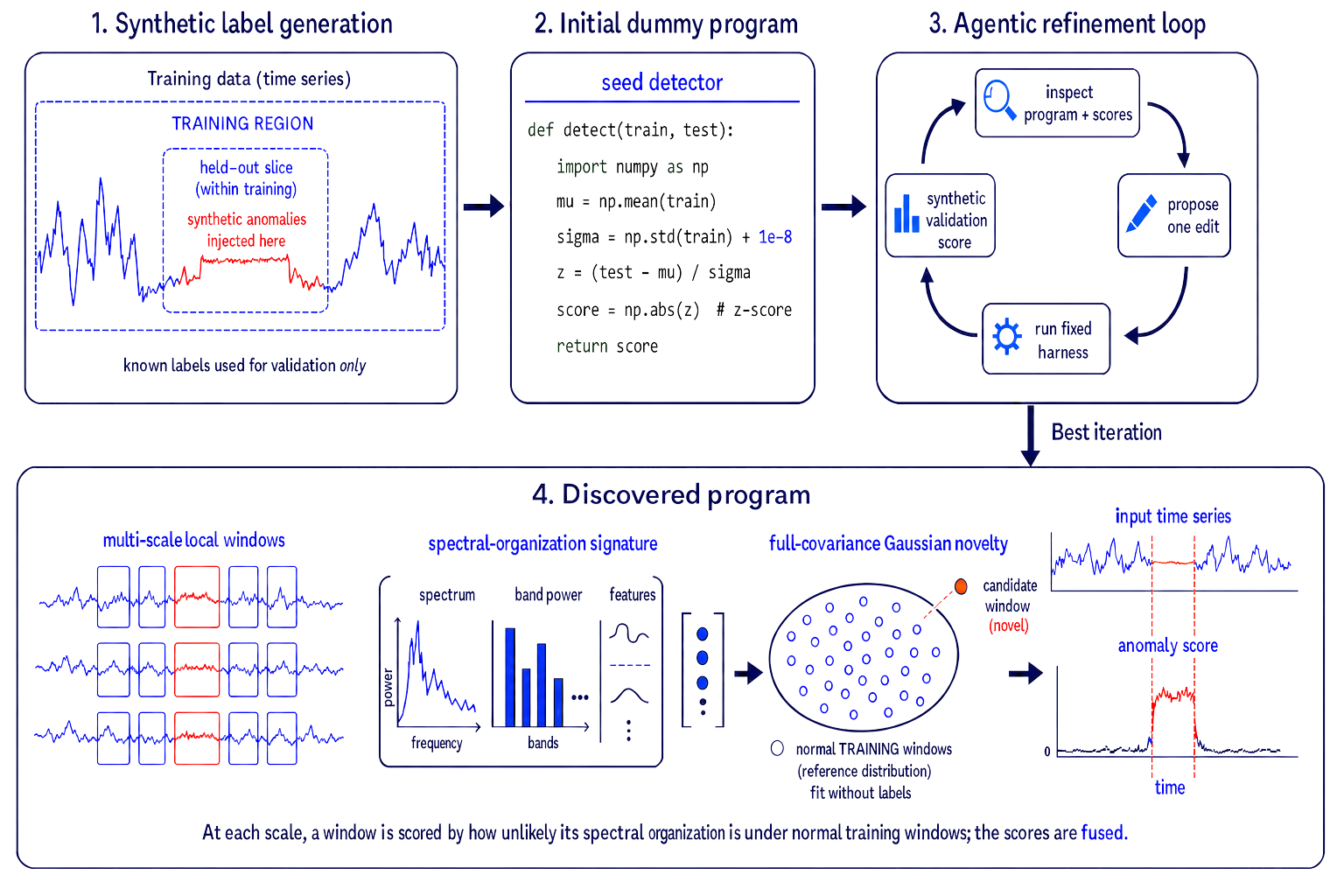}
  \caption{Autonomous discovery of an unsupervised anomaly detector.
  Synthetic anomalies are injected and labeled only in a held-out slice of the training region, and these labels are used only for validation; no real labels or official-test-suffix data enter the search.
  Starting from a simple seed detector, the agent repeatedly proposes program edits and evaluates it with the fixed harness.
  The best iteration yields a multi-scale spectral-organization detector: each window is compared with a full-covariance Gaussian fit to normal training windows, and its per-scale novelty scores are fused.}
  \label{fig:autonomous-loop}
\end{figure*}

\section{Related work}
\label{sec:related}

\paragraph{Benchmarks and evaluation for time series anomaly detection}
Our primary benchmark is TSB-AD~\citep{liuelephant}, which we evaluate under the 16-family protocol published with Time-RCD~\citep{lan2026timercd}.
On TSB-AD we compare against detectors spanning pretrained, deep, and classical families: the recent Time-RCD foundation model~\citep{lan2026timercd}; the deep reconstruction and forecasting detectors TranAD~\citep{tuli2022tranad}, USAD~\citep{audibert2020usad}, OmniAnomaly~\citep{su2019robust}, and DCdetector~\citep{yang2023dcdetector}, together with the temporal-frequency masked autoencoder TFMAE~\citep{fang2024tfmae}; and the classical baselines LOF~\citep{breunig2000lof}, Isolation Forest~\citep{liu2008isolation}, and subsequence PCA~\citep{shyu2003novel}.
For a broader picture we additionally report results on the more recent TAB benchmark~\citep{qiu2025tab} in Appendix~\ref{app:full}.

\paragraph{Autonomous research agents and LLM-driven discovery}
A growing body of work uses large language models not as the final model but as an engine for proposing and refining code.
FunSearch showed that searching over LLM-proposed programs can yield new results in mathematics~\citep{romeraparedes2023mathematical}; AlphaEvolve broadened this into a general coding agent for scientific and algorithmic discovery~\citep{novikov2025alphaevolve}; and open frameworks such as ShinkaEvolve study how to make such program search sample-efficient~\citep{lange2025shinka}.
Most directly, \citet{karpathy2026autoresearch} articulates the idea of an autonomous-research loop in a CLI in which an LLM agent repeatedly proposes an edit to a single artifact, evaluates it against a fixed metric, and iterates without human intervention.
Closest in application, EVIL evolves compact, interpretable algorithms for zero-shot inference on event sequences and time series~\citep{berghaus2026evil}.
This program-authoring role is distinct from using a language model \emph{as} the time-series model: One-Fits-All repurposes a frozen pretrained language-model backbone for downstream tasks including anomaly detection~\citep{gpt4ts}, and time-series foundation models score data directly from learned weights~\citep{lan2026timercd}.
In our setting the language model never sees the data at inference; it only writes the detector program, and the object that scores each series is a few lines of NumPy.

\paragraph{Synthetic data for foundation models}
Synthetic simulation is increasingly used to pretrain foundation models for time series and event sequences, where it supplies a broad prior over mechanisms that can be adapted or applied zero-shot to real data.
% The Foundation Inference Model (FIM) line pretrains on simulated process families for temporal point-process inference, time-series imputation, Markov jump processes, and ordinary and stochastic differential equations~\citep{fim_pp,seifner2025zeroshot,fim_mjp,fim_ode,fim_sde}.
Related time-series foundation-model work uses synthetic anomalies and relative contexts (Time-RCD), causally coherent kernel-generated series (CauKer), or synthetic series paired with symbolic expressions (SymTime)~\citep{lan2026timercd,cauker2025,symtime2025}.

\section{Discovering detectors with an autonomous research loop}
\label{sec:method}

\paragraph{Detector representation}
Each candidate detector is a self-contained Python and NumPy function with a fixed signature: it consumes a training segment and a test segment of a series and returns a per-timestep anomaly score.
Restricting candidates to short array programs has two benefits.
The programs are cheap to run, so a single evaluation across the validation set takes minutes rather than hours, which is what makes an iterative loop practical; and they remain interpretable, so the design choices that perform well can be read directly from the code.

\paragraph{The loop}
We use an autonomous research loop similar to~\citep{karpathy2026autoresearch}: an LLM-driven agent that edits and re-evaluates a single program with no human intervention.
A fixed harness defines the data interface, the validation objective, and the scoring code. The language model is allowed to edit only the detector program.
In each iteration it (i)~inspects the current detector program and its recent scores, (ii)~proposes a concrete modification, ranging from a hyperparameter change to a new scoring component or a structural rewrite, (iii)~runs the modified program through the harness on the validation split, and (iv)~reads the resulting validation score.
Unlike a hill-climbing search, the loop applies no automatic accept/reject step: the full trajectory of attempts, including edits that lower the score, stays in the agent's context and serves as inspiration for later proposals, and the single program returned by the search is the best-scoring one over the whole session (formalized below).
We steer the loop toward short, single-principle programs rather than accuracy alone, which keeps the discovered detectors readable instead of letting them grow into opaque ensembles.
We run each search independently for the univariate and multivariate settings and with several driving models (Section~\ref{sec:setup}).
The strongest loop, driven by Opus 4.8, discovers related spectral-Gaussian novelty patterns in the two regimes, whereas the other loops and the evolutionary baseline arrive at different programs (Section~\ref{sec:shinka}).

\paragraph{Synthetic validation and leakage control}
The search uses \emph{no real anomaly labels}.
Candidates are scored by detecting synthetic anomalies injected only into a held-out slice of each series' \emph{training} region.
The generator is an explicit prior over what an anomaly can look like. We aim for anomalies that are realistic in the mechanism that produces them, like a sensor that freezes, a valid shape appearing out of context, a cycle at the wrong period or phase.
Its types are taken from the established GutenTAG taxonomy~\citep{wenig2022timeeval}, and its mix spans both the magnitude and the in-range shape/contextual families, so that a candidate cannot score well by handling only one of them.
The univariate mix emphasizes the in-range families, which cannot be caught from magnitude alone, while the multivariate mix also includes joint cross-channel perturbations.
Appendix~\ref{app:search} gives the complete generator and leakage-control protocol.

\section{Methods discovered by the searches}
\label{sec:detectors}

Table~\ref{tab:discovered-methods} summarizes the programs selected by all four searches.
The strongest search, the Opus~4.8 loop, found two spectral-Gaussian novelty methods and performed best overall; we therefore describe those two methods below.
The other loops produced simpler local-feature methods, while the Shinka evolutions produced ensembles of standard signals.
The evaluations can be found in Section~\ref{sec:results}; constants and complete executable implementations of the Opus methods are in Appendices~\ref{app:hyper} and~\ref{app:algorithms}.
The complete, runnable source code of every discovered detector (all four searches, in both the univariate and multivariate regimes) is available in an anonymized repository.\footnote{\url{https://anonymous.4open.science/r/TS-AD_submission-7E42/README.md}}

\begin{table*}[t]
  \centering
  \caption{Methods selected by each search.
  Entries summarize the final program in each regime; full accuracy results appear in Section~\ref{sec:results}.}
  \label{tab:discovered-methods}
  \small
  \setlength{\tabcolsep}{5pt}
  \begin{tabular}{p{0.18\textwidth}p{0.35\textwidth}p{0.35\textwidth}}
    \toprule
    Search & Univariate method & Multivariate method \\
    \midrule
    Opus 4.8 Loop & Multi-scale spectral-organization signature scored by Gaussian novelty & Joint per-channel band-powers scored by Gaussian novelty \\
    GPT 5.6 Terra Loop & Local total variation in training-percentile coordinates & Local distribution-shape field \\
    MiniMax-2.5 Loop & Local derivative and higher-moment feature fusion & Z-score, interpolation-error, and local-variance fusion \\
    MiniMax-2.5 ShinkaEvolve & Ensemble of autocorrelation, entropy, divergence, rank, and z-score signals & Ensemble of Mahalanobis, autoregressive-residual, z-score, and derivative signals \\
    \bottomrule
  \end{tabular}
\end{table*}

\paragraph{The Opus methods}
The Opus loop was run once for each regime, and both times it arrived at a detector built on the same idea.
Each one slides a short window along the series, describes the rhythms inside that window, and then asks how unusual that description looks next to normal stretches of the training data.
Windows that stand out get a high anomaly score, which is turned back into a per-timestamp score.
Both detectors are fast, with a running time that grows roughly in proportion to the length of the series.
Where they differ is the description they use: the univariate detector captures how a window's energy is spread across its rhythms, while the multivariate detector captures the combined rhythms of all channels at once.

\paragraph{The univariate detector}
Rather than looking at the raw values in a window, this detector looks at how the window's energy is split between fast and slow rhythms, and at the overall shape of that split.
It turns each window into a short list of descriptive numbers, for example how peaked or how flat the spectrum is, where its energy sits, and how much of it falls in slow versus fast rhythms, together with a term that reacts when part of the window suddenly goes quiet or freezes.
Earlier frequency-based detectors compare a spectrum against a smoothed baseline~\citep{ren2019time}, score windows by their distance to a nearest neighbour~\citep{yeh2016matrix}, or learn a frequency-domain reconstruction~\citep{fang2024tfmae,xu2024fitsmodelingtimeseries}; what is different here is that the description is fixed, easy to read, and judged as a whole rather than one number at a time.
The score therefore rises not just when a single number is unusual, but when the normal relationship between them breaks.
Because the description is mostly about shape and rhythm rather than the raw size of the values, the detector also catches anomalies whose values stay in the normal range and that value-based detectors tend to miss, such as cycles running at the wrong period or phase, out-of-context subsequences, and frozen segments.
It repeats this at several window sizes and blends the results, giving extra weight to the windows that look genuinely improbable.
To our knowledge this particular construction has not previously been used for anomaly detection, and we verify empirically that it outperforms the closest existing frequency-domain and feature-based detectors (Section~\ref{sec:closest}).

Formally, a window $x\in\mathbb{R}^{w}$ is Hann-tapered and mean-removed, $u=(x-\bar{x})\odot h$, with power spectrum $P_k=|\mathrm{FFT}(u)_k|^{2}$, $k=1,\dots,m$ ($m=\lfloor w/2\rfloor$), normalized to $p_k=P_k/\sum_j P_j$ ($\hat{\sigma}$ a standard deviation, $\delta$ a small constant).
Its spectral-organization signature stacks eight scalars,
\begin{equation}
\begin{aligned}
  H&=-\textstyle\sum_k p_k\log p_k, & \kappa&=\max\nolimits_k p_k,\\
  c&=\textstyle\sum_k k\,p_k, & \varepsilon_E&=\log(\hat{\sigma}_x+\delta),\\
  \mathrm{SF}&=\log\tfrac{(\prod_k P_k)^{1/m}}{m^{-1}\sum_k P_k}, & \rho&=\log\tfrac{\sum_{k\le m/2}P_k}{\sum_{k>m/2}P_k},\\
  \beta&=\tfrac{\sum_k(\ell_k-\bar{\ell})\log P_k}{\sum_k(\ell_k-\bar{\ell})^2}, & \gamma&=\log\tfrac{\min_b\hat{\sigma}_{x^{[b]}}}{\hat{\sigma}_x},
\end{aligned}
\label{eq:csosn-sig}
\end{equation}
namely the entropy $H$, peak concentration $\kappa$, centroid $c$, log-energy $\varepsilon_E$, log flatness $\mathrm{SF}$, low/high band ratio $\rho$, log--log spectral slope $\beta$ ($\ell_k=\log k$), and a complexity-collapse term $\gamma$ over four sub-blocks $x^{[b]}$.
Appending a bank of $N=11$ log-spaced band-powers $\phi_{\mathrm{band},b}=\log(\sum_{k\in\mathcal{B}_b}p_k+\delta)$ gives the descriptor $\phi(x)\in\mathbb{R}^{19}$.
This descriptor is scored by a shrunk full-covariance Mahalanobis distance to the Gaussian of normal windows and fused across a dyadic scale space with a tail-emphasized weighting. The exact score and fusion, and all constants, are given in Appendix~\ref{app:hyper} (Eqs.~\ref{eq:csosn-maha}--\ref{eq:csosn-fusion}).

\paragraph{The multivariate detector}
For series with many channels, the detector describes the rhythms of all channels together.
It slides a window across the channels, removes each window's level so that only its shape matters, and measures how much energy each channel carries in a few frequency bands.
Stacking these across all channels gives one description of the window's joint rhythm.
It then builds a single statistical model of what normal windows look like and scores each window by how far it lies from that model.
Because all channels share one model, the score reacts not only when a single channel's rhythm changes, but when the usual relationship between the channels breaks, even if every channel on its own still looks perfectly normal.
To stay fast, the detector samples windows at regular strides and caps how many it uses, so its cost grows roughly in proportion to the length of the series even for very long, high-dimensional data.
And because the model has a fixed size rather than growing with the data, one fit on a short training segment carries over to much longer test series.
The full description and all constants are given in Appendix~\ref{app:hyper} (Eqs.~\ref{eq:xscn-desc}--\ref{eq:xscn-maha}).

\section{Experimental setup}
\label{sec:setup}

\paragraph{Benchmarks and datasets}
Our primary evaluation is the established TSB-AD~\citep{liuelephant}, using the 16-family protocol of Time-RCD~\citep{lan2026timercd}: eleven univariate families (IOPS, MGAB, NAB, NEK, Power, SED, Stock, TODS, UCR, WSD, Yahoo) and five multivariate families (MSL, PSM, SMAP, SMD, SWaT), all listed in Table~\ref{tab:timercd-protocol-full}.
For completeness we additionally evaluate on TAB~\citep{qiu2025tab}, a more recent benchmark that assembles a larger and differently curated field, both univariate ($14$ dataset families, $816$ series) and multivariate ($25$ families, $214$ series), under a single evaluation pipeline and reports a wider panel of range- and volume-aware metrics. We report its full fields in Appendix~\ref{app:full}.

\paragraph{Metrics}
On TSB-AD we follow the Time-RCD protocol and report the affiliation F-measure~\citep{huet2022local}, a tolerance-aware temporal F1 (F1-T), the point-wise standard F1, and VUS-PR~\citep{paparrizos2022volume}.
On TAB we report the panel it standardizes: the volume-under-the-surface pair (VUS-PR, VUS-ROC), the two range-aware areas under the curve (R-AUC-PR, R-AUC-ROC)~\citep{paparrizos2022volume}, and the affiliation F-measure~\citep{huet2022local}.
Precise definitions are in Appendix~\ref{app:metrics}.

\paragraph{Baselines and protocol}
On TSB-AD we adopt the published 16-family protocol of Time-RCD~\citep{lan2026timercd} and place our detectors, scored under that protocol, against the baseline results reported for it: the Time-RCD foundation model~\citep{lan2026timercd}, the deep detectors TranAD~\citep{tuli2022tranad}, USAD~\citep{audibert2020usad}, OmniAnomaly~\citep{su2019robust}, DCdetector~\citep{yang2023dcdetector}, and TFMAE~\citep{fang2024tfmae}, and the classical baselines LOF~\citep{breunig2000lof}, Isolation Forest~\citep{liu2008isolation}, and subsequence PCA~\citep{shyu2003novel}.
Time-RCD is the strongest non-classical baseline in this field.
On TAB (Appendix~\ref{app:full}) we compare against the full field of detectors bundled with the benchmark, spanning non-learning, classical, and deep methods (for example HBOS~\citep{goldstein2012histogram}, EIF~\citep{hariri2019extended}, KMeans, DWT-MLEAD~\citep{thill2017time}, and deep models such as TimesNet~\citep{wu2023timesnet}, ModernTCN~\citep{luo2024moderntcn}, PatchTST~\citep{nie2023patchtst}, iTransformer~\citep{liu2023itransformer}, CATCH~\citep{wu2024catch}, DUET~\citep{qiu2025duet}, and the Anomaly Transformer~\citep{xu2022anomaly}), together with the foundation models Time-RCD~\citep{lan2026timercd}, TSPulse~\citep{ekambaram2025tspulse}, and Chronos~\citep{ansari2024chronos} (the last scored by its one-step forecast residuals, univariate only), all run from their released checkpoints without tuning.
We verified that our evaluation scripts reproduced the numbers reported in the paper, confirming the wrappers are faithful and that the comparison is one-to-one.

\paragraph{The searches we compare}
As ablations we ask how much the discovered detector depends on the driving model and on the search procedure, rather than on the harness alone.
We compare four searches that hold the harness, the seed program, the synthetic-validation objective, and the single-principle prompt \emph{fixed}, varying only the model and the search procedure.
Three are instances of the agentic loop above, driven by various language models (Claude Opus 4.8, GPT 5.6 Terra, and the smaller open MiniMax-2.5), each run at the model's \emph{default} reasoning-effort setting for a single one-hour session of roughly $100$ evolutions per regime.
The fourth swaps the agentic loop for \textsc{ShinkaEvolve}~\citep{lange2025shinka}, an off-the-shelf evolutionary program-search framework (island populations with an elite archive) in the style of AlphaEvolve~\citep{novikov2025alphaevolve}, driven by the same open MiniMax-2.5 model for a matched budget of $100$ generations per regime.
For every search we select the program with the best synthetic-validation score and evaluate it on both benchmarks exactly as we evaluate the baselines.

\section{Results}
\label{sec:results}

\subsection{Accuracy on TSB-AD}

Table~\ref{tab:timercd-protocol-full} reports accuracy on TSB-AD under the Time-RCD protocol, across the eleven univariate and five multivariate dataset families and the four protocol metrics (affiliation F-measure, temporal F1-T, standard F1, and VUS-PR).
The \emph{Opus 4.8 Loop} row applies the univariate detector to the univariate families and the multivariate detector to the multivariate families; within each metric the best value per family is marked with a podium medal (\gmedal{}\,\smedal{}\,\bmedal{} for first, second, and third), and the final three columns count how often each method places first, second, or third across the sixteen families.

\paragraph{The Opus loop leads the field}
The detector discovered by the Opus 4.8 loop takes the most first-place finishes on every one of the four metrics, and its advantage is widest on the precision-oriented metrics (standard F1 and VUS-PR).
Its closest competitor is Time-RCD, a foundation model pretrained for zero-shot anomaly detection. The remaining deep detectors and the classical baselines trail further back and top only isolated families.
Among our own searches the one exception is the multivariate side of TSB-AD, where the GPT~5.6~Terra loop's detector is marginally ahead of the Opus one (Section~\ref{sec:shinka}, Table~\ref{tab:timercd-protocol-variants-full}). The Opus detector remains clearly in front on the univariate families and on both halves of TAB (Appendix~\ref{app:full}, Tables~\ref{tab:tab-uni-full} and~\ref{tab:tab-multi-full}).

\begin{figure}[t]
  \centering
  \includegraphics[width=\columnwidth]{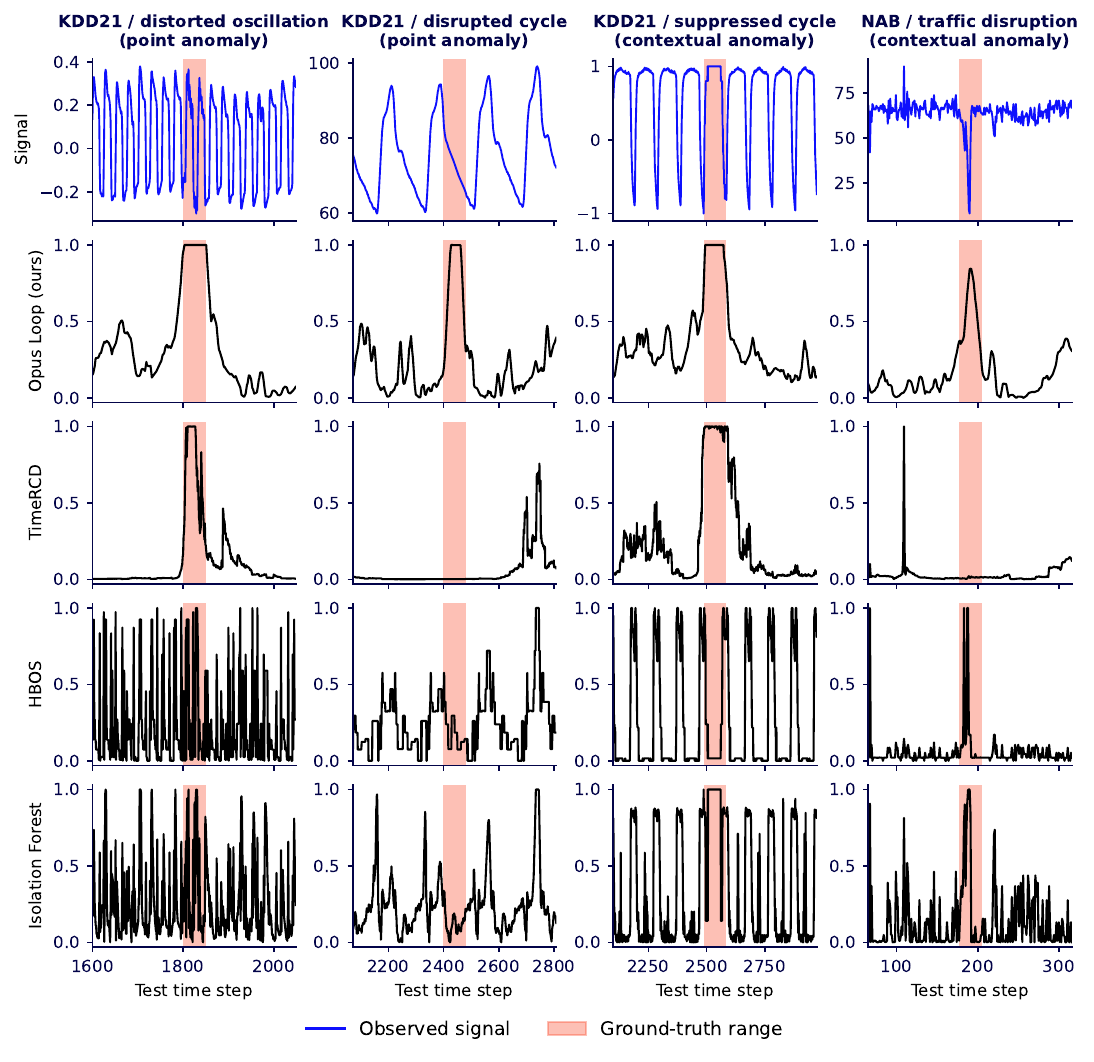}
  \caption{Qualitative score comparison on four representative TAB univariate held-out test suffixes: two point anomalies (left) and two contextual anomalies (right).
  Each column shows the observed signal and the scores from the Opus Loop's discovered detector, Time-RCD~\citep{lan2026timercd}, HBOS~\citep{goldstein2012histogram}, and Isolation Forest~\citep{liu2008isolation}; orange bands mark the labeled anomaly range.
  The Opus Loop produces a sustained high score over each labeled event, while the comparators either miss the event or also respond strongly to normal cycles.
  For display, each score trace is independently normalized to $[0,1]$, preserving its within-series ordering.}
  \label{fig:qualitative-scores}
\end{figure}

\paragraph{Corroboration on TAB}
The larger TAB field (Appendix~\ref{app:full}, Tables~\ref{tab:tab-uni-full} and~\ref{tab:tab-multi-full}) points the same way, with some nuance.
On multivariate TAB the Opus loop's detector is first on all four range-aware metrics, ahead of every classical and deep baseline.
On univariate TAB it leads the ranking-oriented ROC metrics, while KMeans is strongest on the precision metrics, so the two share the top of that field.
On the region-based affiliation F-measure the detector is only mid-field on TAB, a consequence of emitting sharp novelty peaks rather than the broad predicted regions that this measure rewards (Section~\ref{sec:discussion}).

\begin{table*}[t]
    \centering
\caption{Evaluation on TSB-AD~\citep{liuelephant} using the setup of Time-RCD~\citep{lan2026timercd}. The \textbf{Opus 4.8 Loop} row combines its two separately discovered, regime-specific algorithms: the univariate algorithm in the univariate columns and the multivariate algorithm in the multivariate columns. Gold, silver, and bronze markers denote first, second, and third place; the final three columns report their respective counts.}
\label{tab:timercd-protocol-full}
    \resizebox{\textwidth}{!}{%
    \begin{tabular}{l|l|ccccccccccc|ccccc|c|c|c}
        \hline
        \multirow{2}{*}{\textbf{Metric}} & \multirow{2}{*}{\textbf{Model}} & \multicolumn{11}{c|}{\textbf{Univariate Datasets}} & \multicolumn{5}{c|}{\textbf{Multivariate Datasets}} & \textbf{Total} & \textbf{Total} & \textbf{Total} \\
        \cline{3-18}
        & & \textbf{IOPS} & \textbf{MGAB} & \textbf{NAB} & \textbf{NEK} & \textbf{Power} & \textbf{SED} & \textbf{Stock} & \textbf{TODS} & \textbf{UCR} & \textbf{WSD} & \textbf{YAHOO} & \textbf{MSL} & \textbf{PSM} & \textbf{SMAP} & \textbf{SMD} & \textbf{SWaT} & \textbf{1st} & \textbf{2nd} & \textbf{3rd} \\
        \hline
\multirow{10}{*}{Affiliation-F} & \textbf{Opus 4.8 Loop (ours)} & \textbf{\gmedal{} 89.04} & \textbf{\gmedal{} 77.36} & 85.73 & 76.57 & \smedal{} 84.60 & \textbf{\gmedal{} 99.00} & 67.91 & \smedal{} 79.66 & \textbf{\gmedal{} 87.55} & \textbf{\gmedal{} 94.36} & \smedal{} 92.02 & \textbf{\gmedal{} 89.51} & \smedal{} 79.45 & \textbf{\gmedal{} 93.89} & \bmedal{} 88.22 & \textbf{\gmedal{} 77.87} & 08 & 04 & 01 \\
 & Time-RCD & \smedal{} 86.98 & \smedal{} 74.54 & \textbf{\gmedal{} 93.81} & 83.95 & \textbf{\gmedal{} 85.02} & \smedal{} 88.99 & 41.31 & \textbf{\gmedal{} 84.86} & \smedal{} 84.34 & \smedal{} 91.00 & \textbf{\gmedal{} 95.94} & \smedal{} 84.66 & \textbf{\gmedal{} 82.31} & 86.01 & \smedal{} 91.75 & 69.43 & 05 & 07 & 00 \\
 & TranAD & \bmedal{} 83.19 & 67.28 & 90.28 & \bmedal{} 85.02 & 71.56 & 61.03 & 57.94 & 52.76 & 73.31 & \bmedal{} 84.34 & 76.08 & 79.91 & \bmedal{} 73.83 & 87.39 & \textbf{\gmedal{} 92.20} & 75.37 & 01 & 00 & 04 \\
 & USAD & 71.08 & 67.81 & \bmedal{} 91.54 & 71.13 & 76.48 & 55.60 & 35.92 & 47.90 & 76.00 & 65.10 & 53.05 & 81.86 & 57.86 & 87.25 & 85.09 & 75.06 & 00 & 00 & 01 \\
 & OmniAnomaly & 80.32 & 67.35 & \smedal{} 92.35 & \smedal{} 86.30 & \bmedal{} 78.16 & 61.26 & \textbf{\gmedal{} 75.24} & 50.73 & 73.53 & 78.02 & 71.31 & 83.15 & 58.17 & \smedal{} 91.38 & 85.82 & 73.39 & 01 & 03 & 01 \\
 & LOF & 81.06 & 68.44 & 75.75 & 84.74 & 66.76 & 63.85 & \bmedal{} 69.74 & 60.58 & 73.53 & 81.29 & 75.63 & \bmedal{} 84.35 & 61.98 & 63.32 & 64.13 & 56.34 & 00 & 00 & 02 \\
 & IForest & 52.81 & \bmedal{} 68.82 & 39.84 & 71.15 & 0.00 & 70.09 & 0.06 & 44.17 & 50.56 & 41.24 & 33.30 & 63.36 & 63.78 & 59.96 & 69.71 & 0.00 & 00 & 00 & 01 \\
 & Sub-PCA & 75.39 & 66.90 & 89.29 & \textbf{\gmedal{} 97.10} & 71.37 & 67.14 & \smedal{} 70.63 & \bmedal{} 72.75 & \bmedal{} 76.66 & 76.45 & 75.85 & 84.25 & 71.49 & \bmedal{} 90.08 & 85.80 & \smedal{} 76.29 & 01 & 02 & 03 \\
 & DCdetector & 71.83 & 67.91 & 72.21 & 62.31 & 69.75 & \bmedal{} 72.20 & 55.79 & 57.81 & 70.18 & 72.79 & 67.77 & 67.74 & 67.32 & 67.10 & 69.55 & 71.07 & 00 & 00 & 01 \\
 & TFMAE & 78.25 & 67.50 & 75.99 & 76.91 & 70.30 & 68.17 & 56.39 & 62.83 & 70.60 & 80.25 & \bmedal{} 76.87 & 75.70 & 70.07 & 75.36 & 70.85 & \bmedal{} 75.72 & 00 & 00 & 02 \\
        \hline
\multirow{10}{*}{F1-T} & \textbf{Opus 4.8 Loop (ours)} & \smedal{} 50.86 & \textbf{\gmedal{} 29.08} & \bmedal{} 53.59 & 58.00 & \textbf{\gmedal{} 29.61} & \textbf{\gmedal{} 83.71} & 15.48 & \smedal{} 46.59 & \textbf{\gmedal{} 41.36} & \textbf{\gmedal{} 64.25} & \bmedal{} 54.27 & \textbf{\gmedal{} 60.71} & \smedal{} 39.20 & \textbf{\gmedal{} 57.49} & \smedal{} 52.41 & 39.82 & 07 & 04 & 02 \\
 & Time-RCD & \bmedal{} 45.39 & \smedal{} 9.05 & 50.49 & \bmedal{} 76.91 & 23.05 & \smedal{} 60.59 & 18.38 & \textbf{\gmedal{} 68.46} & \smedal{} 41.18 & \smedal{} 44.13 & \textbf{\gmedal{} 79.14} & 43.67 & \textbf{\gmedal{} 43.86} & 35.82 & \textbf{\gmedal{} 55.96} & 35.47 & 04 & 04 & 02 \\
 & TranAD & 22.63 & 1.65 & 37.28 & 69.97 & 22.36 & 9.57 & 16.73 & 13.51 & 7.75 & 20.94 & 8.41 & 39.42 & 25.49 & 29.12 & 37.98 & \smedal{} 49.58 & 00 & 01 & 00 \\
 & USAD & 20.99 & \bmedal{} 4.07 & \textbf{\gmedal{} 61.46} & 70.64 & \smedal{} 28.23 & 9.54 & 16.86 & 20.85 & 14.63 & 14.18 & 9.35 & 48.71 & 28.96 & 43.94 & 50.41 & \textbf{\gmedal{} 50.41} & 02 & 01 & 01 \\
 & OmniAnomaly & \textbf{\gmedal{} 51.17} & 1.61 & 40.09 & \smedal{} 82.20 & \bmedal{} 23.48 & 9.68 & \smedal{} 36.22 & 14.33 & 8.47 & \bmedal{} 34.79 & 24.16 & \smedal{} 49.36 & \bmedal{} 30.42 & \smedal{} 46.63 & 51.84 & 46.64 & 01 & 04 & 03 \\
 & LOF & 27.97 & 1.15 & 35.76 & 63.57 & 19.80 & 9.60 & \textbf{\gmedal{} 66.14} & \bmedal{} 31.63 & 8.31 & 24.38 & \smedal{} 55.93 & 38.97 & 25.58 & 21.81 & 10.13 & 30.62 & 01 & 01 & 01 \\
 & IForest & 7.64 & 0.84 & 21.44 & 65.56 & 0.00 & 9.54 & 1.10 & 11.06 & 6.36 & 4.28 & 4.90 & 20.73 & 25.39 & 14.32 & 16.20 & 0.00 & 00 & 00 & 00 \\
 & Sub-PCA & 32.75 & 0.98 & \smedal{} 54.15 & \textbf{\gmedal{} 84.18} & 20.30 & 9.54 & 20.10 & 18.40 & \bmedal{} 18.90 & 24.62 & 11.57 & \bmedal{} 49.02 & 30.38 & \bmedal{} 44.31 & \bmedal{} 51.87 & \bmedal{} 46.65 & 01 & 01 & 05 \\
 & DCdetector & 6.61 & 1.32 & 32.72 & 29.21 & 21.13 & 10.53 & 16.07 & 16.40 & 6.62 & 7.32 & 6.81 & 23.24 & 25.34 & 15.73 & 9.47 & 28.64 & 00 & 00 & 00 \\
 & TFMAE & 19.41 & 1.07 & 33.04 & 31.11 & 20.18 & \bmedal{} 11.82 & \bmedal{} 22.15 & 16.48 & 5.90 & 19.11 & 23.85 & 25.28 & 25.36 & 19.39 & 10.13 & 28.46 & 00 & 00 & 02 \\
        \hline
\multirow{10}{*}{Standard-F1} & \textbf{Opus 4.8 Loop (ours)} & \textbf{\gmedal{} 47.97} & \textbf{\gmedal{} 28.96} & \bmedal{} 45.09 & 47.43 & \textbf{\gmedal{} 29.64} & \textbf{\gmedal{} 83.74} & 16.91 & \smedal{} 54.26 & \textbf{\gmedal{} 36.56} & \textbf{\gmedal{} 60.54} & \smedal{} 52.33 & \textbf{\gmedal{} 49.01} & \textbf{\gmedal{} 31.01} & \textbf{\gmedal{} 53.10} & 50.79 & 49.19 & 09 & 02 & 01 \\
 & Time-RCD & \smedal{} 47.28 & \smedal{} 8.63 & 42.66 & \bmedal{} 68.64 & 23.12 & \smedal{} 60.85 & 19.56 & \textbf{\gmedal{} 70.02} & \smedal{} 33.47 & \smedal{} 41.04 & \textbf{\gmedal{} 89.05} & 34.48 & 28.25 & 25.36 & \bmedal{} 53.82 & 34.44 & 02 & 05 & 02 \\
 & TranAD & 34.85 & 1.46 & 27.33 & 60.36 & 22.36 & 2.63 & 16.23 & 11.94 & 4.40 & 20.23 & 5.70 & 29.60 & 25.63 & 25.11 & 43.99 & \smedal{} 61.86 & 00 & 01 & 00 \\
 & USAD & 30.66 & \bmedal{} 3.89 & \textbf{\gmedal{} 56.15} & 62.91 & \smedal{} 28.24 & 3.41 & 17.99 & 23.87 & 10.74 & 13.20 & 7.21 & \bmedal{} 38.71 & 28.41 & 38.66 & 53.06 & \textbf{\gmedal{} 62.82} & 02 & 01 & 02 \\
 & OmniAnomaly & \bmedal{} 47.05 & 1.44 & 28.81 & \smedal{} 74.03 & \bmedal{} 23.50 & 0.43 & \smedal{} 38.59 & 12.65 & 5.11 & \bmedal{} 29.57 & 21.40 & \smedal{} 39.10 & \smedal{} 30.43 & \smedal{} 40.50 & \smedal{} 57.06 & 55.93 & 00 & 06 & 03 \\
 & LOF & 30.28 & 1.05 & 24.04 & 56.92 & 12.18 & 4.11 & \textbf{\gmedal{} 66.20} & \bmedal{} 25.77 & 4.70 & 22.62 & \bmedal{} 48.95 & 30.65 & 18.80 & 18.70 & 8.41 & 29.08 & 01 & 00 & 02 \\
 & IForest & 8.37 & 0.73 & 29.41 & 58.10 & 19.77 & 3.81 & 16.91 & 13.35 & 4.09 & 2.07 & 3.20 & 14.68 & 24.15 & 13.61 & 16.89 & 26.76 & 00 & 00 & 00 \\
 & Sub-PCA & 33.96 & 0.83 & \smedal{} 46.71 & \textbf{\gmedal{} 85.43} & 16.05 & 9.56 & 21.77 & 18.72 & \bmedal{} 15.12 & 24.74 & 11.06 & 38.29 & \bmedal{} 30.26 & \bmedal{} 38.90 & \textbf{\gmedal{} 57.22} & \bmedal{} 55.98 & 02 & 01 & 04 \\
 & DCdetector & 5.19 & 1.21 & 24.02 & 17.37 & 21.10 & 10.54 & 16.97 & 17.85 & 3.18 & 4.64 & 4.14 & 14.08 & 25.33 & 10.67 & 8.99 & 27.02 & 00 & 00 & 00 \\
 & TFMAE & 9.48 & 0.97 & 23.78 & 19.74 & 20.14 & \bmedal{} 11.87 & \bmedal{} 23.32 & 14.88 & 2.83 & 15.53 & 20.50 & 15.68 & 25.39 & 12.58 & 9.16 & 27.08 & 00 & 00 & 02 \\
        \hline
\multirow{10}{*}{VUS-PR} & \textbf{Opus 4.8 Loop (ours)} & \textbf{\gmedal{} 49.52} & \textbf{\gmedal{} 14.99} & 41.45 & 46.88 & \textbf{\gmedal{} 21.47} & \textbf{\gmedal{} 97.46} & 66.70 & \smedal{} 84.71 & \textbf{\gmedal{} 33.09} & \textbf{\gmedal{} 60.93} & \smedal{} 75.20 & \textbf{\gmedal{} 40.51} & \textbf{\gmedal{} 22.15} & \textbf{\gmedal{} 47.06} & 36.01 & 30.08 & 09 & 02 & 00 \\
 & Time-RCD & \smedal{} 42.98 & \smedal{} 3.09 & \bmedal{} 44.65 & \bmedal{} 71.13 & \bmedal{} 14.63 & \smedal{} 69.86 & 72.73 & \textbf{\gmedal{} 91.67} & \smedal{} 27.48 & \smedal{} 29.95 & \textbf{\gmedal{} 87.88} & 23.41 & \smedal{} 21.69 & 19.68 & \textbf{\gmedal{} 46.70} & 18.47 & 03 & 06 & 03 \\
 & TranAD & 21.61 & 0.64 & 24.82 & 61.63 & 13.04 & 5.75 & 78.08 & 47.33 & 2.25 & 12.20 & 25.78 & 14.78 & 16.49 & 13.37 & 28.34 & \textbf{\gmedal{} 47.37} & 01 & 00 & 00 \\
 & USAD & 16.58 & \bmedal{} 0.75 & \textbf{\gmedal{} 55.03} & 58.53 & \smedal{} 18.68 & 4.37 & 74.53 & \bmedal{} 56.36 & 8.85 & 10.00 & 14.15 & 29.95 & 17.59 & 26.37 & 34.53 & \smedal{} 44.73 & 01 & 02 & 02 \\
 & OmniAnomaly & \bmedal{} 25.35 & 0.64 & 27.17 & \smedal{} 74.51 & 14.32 & 6.20 & \textbf{\gmedal{} 91.29} & 45.55 & 2.40 & 16.37 & 29.26 & \smedal{} 31.57 & \bmedal{} 18.58 & \smedal{} 28.07 & \bmedal{} 37.44 & 42.97 & 01 & 03 & 03 \\
 & LOF & 19.43 & 0.57 & 21.18 & 58.52 & 9.31 & 6.81 & \smedal{} 83.07 & 49.14 & 2.39 & 12.85 & \bmedal{} 41.37 & 24.67 & 13.58 & 10.59 & 4.40 & 14.50 & 00 & 01 & 01 \\
 & IForest & 8.59 & 0.62 & 23.57 & 56.50 & 11.56 & 7.71 & 70.99 & 46.62 & 2.88 & 2.06 & 10.47 & 11.29 & 15.85 & 7.55 & 8.88 & 15.49 & 00 & 00 & 00 \\
 & Sub-PCA & 23.02 & 0.60 & \smedal{} 46.08 & \textbf{\gmedal{} 88.91} & 10.49 & 3.72 & \bmedal{} 80.86 & 54.16 & \bmedal{} 12.92 & \bmedal{} 16.41 & 21.57 & \bmedal{} 31.43 & 18.52 & \bmedal{} 26.42 & \smedal{} 37.50 & \bmedal{} 43.02 & 01 & 02 & 06 \\
 & DCdetector & 5.83 & 0.59 & 16.60 & 14.03 & 12.32 & 9.37 & 74.16 & 46.66 & 1.53 & 3.23 & 10.17 & 7.01 & 14.49 & 4.21 & 4.66 & 15.04 & 00 & 00 & 00 \\
 & TFMAE & 5.32 & 0.64 & 15.68 & 17.81 & 11.90 & \bmedal{} 9.55 & 73.54 & 48.79 & 2.57 & 5.36 & 25.93 & 8.25 & 14.22 & 5.76 & 4.77 & 15.38 & 00 & 00 & 01 \\
        \hline
    \end{tabular}
    }
    \vspace{-10pt}
\end{table*}

\subsection{Did the loop rediscover an existing detector?}
\label{sec:closest}

The discovered detectors describe each window by a fixed feature vector and score it by its distance to a model of normal windows, a design that overlaps with two established families: \emph{frequency-domain} detectors and \emph{feature-based} subsequence detectors.
It is therefore fair to ask whether the loop merely rediscovered a method that already exists.
To test this directly, we compare against the published methods closest to this design under the identical protocol, each run from its released implementation (Table~\ref{tab:closest}).
Two are canonical frequency-domain detectors: spectral-residual saliency~\citep{ren2019time} and a Fourier-reconstruction detector~\citep{rasheed2009fft}.
The others build on catch22~\citep{lubba2019catch22}, the canonical interpretable time-series feature set, which we use in two forms.
The first is C22MP~\citep{tafazoli2023c22mp}, which scores catch22 features by a matrix profile.
The second, our sharpest control, pairs the catch22 features with our \emph{own} novelty framework: they are scored by the same Mahalanobis distance to a Gaussian of normal windows (``catch22\,+\,Mahalanobis'').

The discovered detector leads all four metrics, and by a wide margin on the precision-oriented VUS-PR ($53.9$ versus at most $35.0$).
The ``catch22\,+\,Mahalanobis'' control is the most telling: holding our novelty framework fixed and substituting the canonical catch22 features for the loop's spectral-organization signature drops VUS-PR from $53.9$ to $28.4$.
The advantage therefore comes from the particular descriptor the loop assembled, not from the feature-based-novelty framework alone; and the discovered detector also clearly outperforms both canonical frequency-domain detectors and the published catch22-based C22MP.
Extending these univariate methods to the multivariate families by scoring each channel independently leaves the ordering unchanged, with the discovered multivariate detector ahead on VUS-PR (Appendix~\ref{app:closest-mv}).

Every ingredient the loop used is individually well established: sliding-window spectral features, a Mahalanobis distance to a Gaussian of normal windows, and multi-scale fusion.
Each of the methods above draws on some of them.
Yet none combines them as the discovered detector does, and the discovered combination measurably outperforms each of the closest existing methods.
We read this as evidence that the loop did not simply rediscover an existing detector, but instead composed known concepts into a new, and better, one.

\begin{table}[t]
  \centering
  \caption{Comparison with the closest published methods on the TSB-AD univariate families (Time-RCD protocol; dataset-mean, higher is better $\uparrow$).
  Each baseline is run from its released implementation and rows are ordered by VUS-PR; ``catch22\,+\,Mahalanobis'' keeps our novelty framework and only substitutes the catch22 feature set for the discovered spectral signature.
  Best per column in \textbf{bold}.}
  \label{tab:closest}
  \small
  \setlength{\tabcolsep}{5pt}
  \begin{tabular}{lcccc}
    \toprule
    Method & Aff-F\,$\uparrow$ & F1-T\,$\uparrow$ & Std-F1\,$\uparrow$ & VUS-PR\,$\uparrow$ \\
    \midrule
    \textbf{Opus 4.8 Loop (ours)} & \textbf{84.89} & \textbf{47.89} & \textbf{45.77} & \textbf{53.86} \\
FFT detector~\citep{rasheed2009fft} & 78.34 & 27.01 & 26.95 & 35.00 \\
Spectral Residual~\citep{ren2019time} & 83.52 & 41.01 & 36.44 & 34.75 \\
C22MP~\citep{tafazoli2023c22mp} & 77.88 & 26.32 & 24.32 & 32.01 \\
catch22 + Mahalanobis~\citep{lubba2019catch22} & 75.40 & 22.96 & 21.34 & 28.36
 \\
    \bottomrule
  \end{tabular}
\end{table}

\subsection{Efficiency}
\label{sec:speed}

Because each detector uses no GPU and trains no neural network, its cost is predictable and modest.
Table~\ref{tab:speed} reports prediction-only wall-clock under the Time-RCD TSB-AD protocol: we time only each detector's scoring call, excluding data loading and metric computation.
Our detectors and the classical baselines run on a single CPU core. The foundation model Time-RCD and the deep detectors, marked $^{*}$, run on a V100 GPU and are timed as deployed: Time-RCD at inference from its pretrained checkpoint, the deep detectors including their per-series fit.
Our detector is the fastest CPU method in both regimes and the fastest method overall on multivariate.
On the univariate families it scores about $5\,\mu$s per step, on par with LOF and $3$--$16\times$ faster than the other classical baselines. Only Time-RCD is faster, and only because a pretrained transformer runs a single batched GPU forward pass, whereas every per-series-trained deep detector is $40$--$120\times$ slower than our detector even on the GPU.
On the multivariate families our detector leads every method: Time-RCD costs about $1.9$~s per series and the deep detectors $2.5$--$7$~s, all on a GPU and all one to two orders of magnitude slower than our CPU-only program, whose classical subsequence competitors also slow steeply on the $25$--$66$-channel series.
In short, the discovered detectors combine their accuracy with runtime and memory that are easy to reason about, as short auditable programs rather than models with GPU-scale cost.

\begin{table}[t]
  \centering
  \caption{Efficiency on TSB-AD (Time-RCD protocol).
  Prediction-only wall-clock (lower is better, $\downarrow$), timing only each detector's scoring call over a representative subset of series (up to five per family, $\le$50k steps); s/series is the per-series mean, $\mu$s/step normalizes by series length.
  Rows marked $^{*}$ run on a V100 GPU; all others use a single CPU core and train no network.}
  \label{tab:speed}
  \footnotesize
  \begin{tabular}{lcc}
    \toprule
    Method & s/series\,$\downarrow$ & $\mu$s/step\,$\downarrow$ \\
    \midrule
    \multicolumn{3}{l}{\emph{Univariate} (44 series)} \\
    Time-RCD~\citep{lan2026timercd}$^{*}$ & \textbf{0.011} & \textbf{0.9} \\
\textbf{Opus 4.8 Loop (ours)} & 0.059 & 5.1 \\
LOF~\citep{breunig2000lof} & 0.068 & 5.9 \\
PCA (subseq.)~\citep{shyu2003novel} & 0.183 & 16.0 \\
Isolation Forest~\citep{liu2008isolation} & 0.575 & 50.2 \\
KMeans & 0.926 & 80.8 \\
USAD~\citep{audibert2020usad}$^{*}$ & 2.2 & 188 \\
TranAD~\citep{tuli2022tranad}$^{*}$ & 6.8 & 596 \\
OmniAnomaly~\citep{su2019robust}$^{*}$ & 6.8 & 597
 \\
    \midrule
    \multicolumn{3}{l}{\emph{Multivariate} (16 series)} \\
    \textbf{Opus 4.8 Loop (ours)} & \textbf{0.054} & \textbf{4.4} \\
LOF~\citep{breunig2000lof} & 1.3 & 104 \\
Time-RCD~\citep{lan2026timercd}$^{*}$ & 1.9 & 152 \\
USAD~\citep{audibert2020usad}$^{*}$ & 2.5 & 209 \\
KMeans & 4.0 & 332 \\
OmniAnomaly~\citep{su2019robust}$^{*}$ & 4.1 & 336 \\
TranAD~\citep{tuli2022tranad}$^{*}$ & 7.3 & 604 \\
PCA~\citep{shyu2003novel} & 29.8 & 2453 \\
Isolation Forest~\citep{liu2008isolation} & 44.8 & 3694
 \\
    \bottomrule
  \end{tabular}
\end{table}

\subsection{Ablation: the model and the search procedure}
\label{sec:shinka}

Holding the harness, the seed program, and the single-principle prompt fixed, the four searches differ only in the driving model and the search procedure, which lets us gauge how much each contributes.
Table~\ref{tab:timercd-protocol-variants-full} compares the four discovered variants on TSB-AD, and Appendix~\ref{app:full} reports the same four in the full TAB field.

\paragraph{The driving model matters most}
The Opus~4.8 loop is the strongest search overall, and by a wide margin in the univariate regime, where it leads the other three on every metric while the other loops (GPT~5.6~Terra and MiniMax-2.5) and the evolutionary baseline cluster well below it.
On the multivariate families the strongest searches sit closer together: the GPT loop edges the Opus loop on TSB-AD, while the Opus loop is clearly ahead on TAB, so across benchmarks neither is consistently in front.

\paragraph{The search procedure matters less, but shapes what gets built}
Fixing the model to MiniMax-2.5 and varying only the search procedure (the agentic loop versus the \textsc{ShinkaEvolve} evolutionary search) gives no consistent winner on accuracy: neither is reliably ahead of the other across the two benchmarks and the two regimes (Table~\ref{tab:timercd-protocol-variants-full} and Appendix~\ref{app:full}), so the search structure looks like a smaller factor than the driving model.
Where the two procedures pull apart is in the kind of program they return.
The agentic loops stayed close to the single-principle steer and produced shorter programs built mainly around one idea: the Opus loop returned two compact, genuinely novel single-principle detectors (Section~\ref{sec:detectors}), and even the other loops kept to a single idea, as with the GPT loop's univariate detector that scores the local total variation of a window in training-percentile coordinates, though those programs transferred to the held-out field far less well, most starkly on univariate.
The \textsc{ShinkaEvolve} search, under the identical prompt asking for one novel principle and explicitly discouraging the fusion of standard methods, instead converged in both regimes on a hacky weighted ensemble of standard methods: several standard signals in the univariate program (autocorrelation structure, gradient-direction entropy, a local distribution-divergence statistic, a local-rank score, and the original z-score signal) and, in the multivariate one, a mix that includes a Mahalanobis distance, per-channel residuals from predicting each channel from the others, the z-score, and velocity and acceleration terms.
We suspect this might be caused by how the framework works: the evolutionary framework often asks to merge different ideas together, and (by construction) it also presents diverse programs as context at each evolution step, so incremental additions to an already competitive program tend to be kept even when they grow a larger, less coherent ensemble rather than a fresh single principle.
A single run per regime cannot pin down that mechanism, but it is a plausible reason the shared prompt did not prevent this kind of heuristic accretion.
The evolutionary programs also end up markedly longer and more heavily hand-tuned than the loops', the evolutionary search is much slower to run at several hours per regime against the loops' roughly one-hour sessions, and its multivariate ensemble is comparatively slow at inference on the longest series, unlike the tight runtime budget the loop's detectors respect.

\section{Discussion and limitations}
\label{sec:discussion}

The appeal of letting an agent discover detectors lies not only in the measured numbers but in the resulting artifact: each detector is a short, readable program, so its inductive biases are inspectable and its cost is predictable.
This contrasts with opaque trained models and makes the method easy to audit and to port.
It also changes where human effort goes, from designing the detector to designing the harness, the synthetic-anomaly generator, and the objective that the agent optimizes against.

Several caveats apply.
First, the detector's standing depends on the metric family and the benchmark.
On TSB-AD it leads across the reported metrics. On TAB it dominates the multivariate side but shares the top of the univariate side with KMeans, which is ahead on the precision metrics there, and it is only mid-field on the region-based affiliation measure because it is a sharp scorer rather than a region predictor.
Second, although we evaluate on two independently curated benchmarks, a still broader study across more benchmarks and dataset families (and potentially real-world deployment) would be needed to confirm the generality of the discovered detectors.
Third, the synthetic generator is where prior knowledge about anomalies enters. We tried to keep it broad and realistic but certain types of anomalies that are not captured well by the prior might be missed by the discovered detectors. The generator is also a design choice that could be varied, and we do not claim that the discovered detectors are optimal for every conceivable anomaly type.
Finally, the discovery has a one-time cost in language-model queries and compute; our efficiency claims concern the discovered detector at inference time, not the search that produced it. Though we do report that the searches could be performed within the 20 USD subscription plans and should therefore be accessible to many researchers and practitioners.

\section{Conclusion}
\label{sec:conclusion}

We asked whether an LLM-driven autonomous research loop is an effective tool for \emph{writing} time-series anomaly detectors.
Driven by frontier models and steered toward short, single-principle programs, the loop discovered related spectral-Gaussian novelty detectors in the two regimes: a rich, multi-scale spectral-organization signature for univariate series, and a joint per-channel band-power model for multivariate series.
On the TSB-AD benchmark these detectors lead the reported metrics and remain near the top of the broader TAB field. Both are fully unsupervised, run on a single CPU core with no GPU and no neural training, and consist of short, readable code that keeps them cheap and auditable.

\section*{Generative AI usage}
The research idea, experimental design, and overall structure of this paper were conceived and developed by the human authors.
LLM agents performed the implementation of the code with human oversight.
Generative AI tools additionally assisted with drafting and editing the manuscript text, in every case under human oversight, with all claims, results, and conclusions verified by the authors.
Figure~\ref{fig:autonomous-loop} was produced with a generative image model.

\section*{Funding}
This research has been funded by the Federal Ministry of Education and Research of Germany and the state of North-Rhine Westphalia as part of the Lamarr Institute for Machine Learning and Artificial Intelligence.

\bibliographystyle{ACM-Reference-Format}
\bibliography{kdd2027_conference}

\appendix

\section{Metric definitions}
\label{app:metrics}

We summarize the standardized metrics used in the main text; full definitions are given by the works we cite.
All accuracy metrics lie in $[0,1]$ and are higher-is-better.

\paragraph{Binarization}
Our detectors output only raw per-timestamp scores $a_t$; we never threshold them ourselves.
The region- and point-based measures that need a binary prediction obtain it inside the evaluation harness, identically for every method (ours and the baselines): under the Time-RCD/TSB-AD protocol of the main text the harness reports each threshold-dependent metric at its score-optimal operating point (sweeping the score quantile and keeping the best attainable value), while on TAB the label metrics are averaged over a fixed grid of contamination ratios.
Threshold-free measures (the AUC and VUS families) use the raw scores $a_t$ directly.

\paragraph{Range-aware AUCs (R-AUC-PR, R-AUC-ROC)}
Standard AUC-PR and AUC-ROC treat every timestamp independently and with equal weight, which understates the quality of detectors on extended (subsequence) anomalies.
The range-aware variants add a tolerance region around each labeled anomaly so that scores raised near a true event are credited, and then integrate precision-recall or ROC behavior over that construction~\citep{paparrizos2022volume}.

\paragraph{Volume under the surface (VUS-PR, VUS-ROC)}
VUS removes the dependence on a single tolerance width by sweeping the size of the boundary buffer and integrating the resulting family of PR or ROC curves into a volume~\citep{paparrizos2022volume}.
This yields a threshold-free and buffer-width-free summary that is well suited to subsequence anomalies, and it is the headline metric on the TSB-AD leaderboard~\citep{liuelephant}.
We order the TAB baseline field by VUS-PR for this reason (listing our discovered detectors first), while reporting the full panel of metrics rather than any composite of them.

\paragraph{Affiliation F-measure (Aff-F)}
The affiliation metric is a parameter-free, label-based measure that pairs predicted anomaly events with ground-truth events and scores their temporal affiliation, combining an affiliation precision and recall into an F-measure~\citep{huet2022local}.
It applies uniformly to point and subsequence anomalies.
Because it scores predicted \emph{regions}, threshold-free scorers like ours are evaluated on the binary regions the harness derives from the raw scores as described above.

\paragraph{Point-wise and temporal F1 (Standard-F1, F1-T)}
The standard F1 is the point-wise F-measure between the binarized predictions $\hat{y}$ and the ground-truth labels $y$, computed timestamp by timestamp.
Writing $\mathrm{TP}$, $\mathrm{FP}$, and $\mathrm{FN}$ for the timestamp-wise true positives, false positives, and false negatives, it is the harmonic mean of precision and recall,
\begin{equation}
  \mathrm{F1}=\frac{2\,\mathrm{TP}}{2\,\mathrm{TP}+\mathrm{FP}+\mathrm{FN}}
  =\left(\tfrac{1}{2}\bigl(\mathrm{Prec}^{-1}+\mathrm{Rec}^{-1}\bigr)\right)^{-1},
\end{equation}
with $\mathrm{Prec}=\mathrm{TP}/(\mathrm{TP}+\mathrm{FP})$ and $\mathrm{Rec}=\mathrm{TP}/(\mathrm{TP}+\mathrm{FN})$.
The temporal F1-T relaxes this with a per-series tolerance around each labeled event, so a prediction close to a true anomaly is credited without requiring exact timestamp overlap.
Both are reported under the Time-RCD protocol on TSB-AD~\citep{lan2026timercd}, with the binary predictions $\hat{y}$ formed inside the harness as described above.

\section{Search configuration}
\label{app:search}

\paragraph{Synthetic anomaly generator}
The validation slice is the last $30\%$ of each series' training region (at least $128$ steps); synthetic anomalies are injected only into this slice with a fixed seed.
Anomalies are placed as non-overlapping windows, sampled until a target fraction of the slice is anomalous (subject to a minimum number of windows and a minimum gap between them), with each window's length drawn uniformly from a fixed range.
Each window receives a single anomaly type, drawn uniformly from a mix that deliberately spans both \emph{magnitude} and \emph{in-range shape/contextual} families, so that a competitive detector must handle both; the types reimplement those in the GutenTAG generator~\citep{wenig2022timeeval}.
Each type is intended to imitate a generic mechanism that occurs in practice: a frozen sensor, a valid shape out of context, a stretched or misaligned cycle, a burst of noise, a drifting level.
Magnitudes are expressed in units of the channel's robust standard deviation $\sigma$: a \emph{mean} shift adds $\mathcal{U}(0.5,4)\,\sigma$ with random sign; a \emph{platform} sets the window to its median $+\,\mathcal{U}(-3,3)\,\sigma$; a \emph{trend} adds a linear ramp of total rise up to $\mathcal{U}(0.5,4)\,\sigma$; a \emph{variance} burst adds Gaussian noise of scale $\mathcal{U}(1.5,6)\,\sigma$; an \emph{extremum} is a single-point spike of $\mathcal{U}(6,40)\,\sigma$; an \emph{amplitude} anomaly rescales the window's oscillation about its mean by $\mathcal{U}(1.5,3)$ (or its reciprocal) under a smooth bell envelope; a \emph{discord} splices in a real subsequence copied from a distant part of the \emph{same} series, level-aligned to the local median (a real but out-of-context shape); a \emph{warp} resamples a neighboring stretch into the window (a stretched/compressed, wrong-period cycle); a \emph{phase} shift circularly rotates the window by a large fraction of its length (a misaligned cycle); a \emph{flatline} freezes the window at its starting value (a frozen sensor); and a \emph{scramble} randomly permutes the window's samples in time (its value distribution is unchanged but its local temporal structure is destroyed).
All types except the point extremum label the entire window; the extremum labels only the spike timestamp.
The two regimes use the following generator configurations, which are the ones used for every experiment in the paper.

\begin{itemize}
  \item \textbf{Univariate.} Anomalous fraction $0.02$; window lengths $\mathcal{U}[4,220]$; minimum gap $32$; at least $3$ windows; single-channel.
  Type mix $\{$\emph{discord}, \emph{warp}, \emph{phase}, \emph{flatline}, \emph{scramble}, \emph{mean}, \emph{platform}, \emph{extremum}$\}$, weighted toward the in-range shape/contextual families, which cannot be resolved from magnitude alone and therefore keep the objective from being satisfied by a threshold on deviation, with level and point types retained for coverage.
  \item \textbf{Multivariate.} Anomalous fraction $0.05$; window lengths $\mathcal{U}[8,96]$; minimum gap $16$; at least $3$ windows; a fraction $0.25$ of anomalies are applied \emph{jointly to all channels} (the rest to a single random channel), which creates cross-channel correlation breaks alongside single-channel events.
  Type mix $\{$\emph{mean}, \emph{platform}, \emph{trend}, \emph{extremum}, \emph{amplitude}, \emph{variance}, \emph{discord}, \emph{warp}$\}$.
\end{itemize}

\section{Discovered detector configurations}
\label{app:hyper}

We give the key constants present in the discovered programs; complete executable implementations appear in Appendix~\ref{app:algorithms}.
Both detectors standardize from training-region statistics, slide a window, derive local spectral features, and score their novelty under a Gaussian model of normal training windows.
The univariate detector uses robust median/MAD feature scaling and a square-root Mahalanobis score; the multivariate detector uses per-channel mean/std scaling and a squared Mahalanobis score.

\paragraph{Univariate detector}
The window is Hann-tapered; its descriptor concatenates the spectral-organization signature (spectral entropy, log spectral flatness, spectral centroid, peak concentration, log energy, log-power/log-frequency slope, low/high band ratio), a within-window complexity-collapse term (least-active sub-block variation vs whole-window variation), and a bank of log-spaced band-powers.
The descriptor enters one shrunk full-covariance ($0.95$ full-covariance weight, shrunk toward a scaled identity) Mahalanobis distance (Eq.~\ref{eq:csosn-maha}), so each coordinate is scored conditional on the others.
The novelty is evaluated over a dyadic scale space of window lengths spanning short to long, and the per-scale scores are fused by a tail-emphasized rank fusion amplified by a gate on the robust-$z$ of a window's distance above the normal training-distance tail.
The per-series pass runs in tens of milliseconds on typical series and at most a few seconds on the longest; the window length adapts to the series, and very long inputs are strided to bound cost.
Formally, writing $z(x)$ for the robustly scaled (median/MAD) descriptor of Eq.~\ref{eq:csosn-sig}, the per-scale novelty is the shrunk full-covariance Mahalanobis distance
\begin{equation}
  s_w(x)=\sqrt{z(x)^{\top}\hat{\Sigma}^{-1}z(x)},\qquad
  \hat{\Sigma}=\lambda\,\widehat{\mathrm{Cov}}\bigl(z^{\mathrm{tr}}\bigr)+(1-\lambda)I,
  \label{eq:csosn-maha}
\end{equation}
with $\lambda=0.95$.
Per-scale distances are referenced to the training tail as $e_w$ and fused over the dyadic scale space $\mathcal{W}$,
\begin{equation}
  a(t)=\frac{1}{|\mathcal{W}|}\sum_{w\in\mathcal{W}}
       F_w\!\bigl(e_w(t)\bigr)^{p}\,\bigl(1+L\,\sigma\bigl(e_w(t)-c\bigr)\bigr),
  \label{eq:csosn-fusion}
\end{equation}
with $F_w$ the training CDF, $\sigma$ the logistic, and $(p,L,c)=(6,8,3)$; the exponent $p$ and the gate concentrate score mass on the extreme tail.

\paragraph{Multivariate detector}
Window $W=80$ (shrunk automatically on short series), a small set of frequency bands ($6$) per channel, all $C$ channels stacked into one descriptor after per-window DC removal and Hann tapering.
The normal training-region descriptors are summarized by their mean and one pooled, ridge-regularized covariance (the sample covariance plus a small diagonal ridge $\eta I$), and the raw score is the squared Mahalanobis distance of each window to that Gaussian (Eq.~\ref{eq:xscn-maha}).
Windows are strided and their number is hard-capped (a few thousand) so the strided FFT and the Mahalanobis stay near-linear on series up to hundreds of thousands of steps and dozens of channels.
Formally, each length-$L$ window ($L=80$) is reduced to per-channel powers $P_{k,c}$ aggregated into $n_b=6$ bands, log-compressed and stacked across channels,
\begin{equation}
  \phi_{b,c}=\log\!\Bigl(1+\textstyle\sum_{k\in\mathcal{B}_b}P_{k,c}\Bigr),\qquad
  \phi\in\mathbb{R}^{n_bC},
  \label{eq:xscn-desc}
\end{equation}
and one Gaussian (mean $m$, ridged covariance $\Sigma$) fitted to the normal descriptors gives the squared Mahalanobis score
\begin{equation}
  d(\phi)=(\phi-m)^{\top}\Sigma^{-1}(\phi-m),\qquad
  \Sigma=\widehat{\mathrm{Cov}}\bigl(\phi^{\mathrm{tr}}\bigr)+\eta I,
  \label{eq:xscn-maha}
\end{equation}
with $\eta=10^{-2}$.
Because $\Sigma$ is a \emph{single pooled} $n_bC\times n_bC$ covariance, its cross-channel off-diagonal blocks encode the normal spectral coupling, so a window that breaks it scores high even when every channel is individually in range.

\section{Discovered algorithms}
\label{app:algorithms}

For full transparency we reproduce the complete executable implementations of the two discovered detectors.
They are the programs evaluated throughout the paper, with only harness edit markers and nonessential explanatory comments removed.
Each is a single self-contained scoring function with a few small helpers; it reads a training segment and a test segment of one series and returns a per-timestamp anomaly score, fitting whatever it needs on the training region without labels.
Algorithm~\ref{lst:csosn} is the univariate detector and Algorithm~\ref{lst:xscn} the multivariate detector, both found by the Opus 4.8 loop.
The detectors found by the other three searches (the GPT~5.6~Terra and MiniMax-2.5 loops and the MiniMax-2.5 \textsc{ShinkaEvolve} baseline, each in both regimes) are provided in full in the anonymized repository linked in Section~\ref{sec:detectors}.\footnote{\url{https://anonymous.4open.science/r/TS-AD_submission-7E42/README.md}}

\begin{lstlisting}[style=pycode,caption={The Opus loop's univariate detector.},label={lst:csosn}]
from __future__ import annotations

import numpy as np


_NBANDS = 11      # log-spaced frequency bands whose log-powers enter the signature
_TAIL_P = 6.0    # tail-power sharpening exponent on the per-scale rank curve
_GATE_C = 3.0    # train-referenced magnitude gate centre (robust-z of distance)
_GATE_L = 8.0    # gate amplification strength (additive 1 + L*gate)
_SHRINK = 0.95    # covariance shrinkage toward identity (full-cov weight)


def _as_1d(values: np.ndarray) -> np.ndarray:
    """Return a 1D float series."""
    a = np.asarray(values, dtype=np.float64)
    if a.ndim == 0:
        return a.reshape(1)
    if a.ndim == 1:
        return a
    return a.reshape(a.shape[0], -1)[:, 0]


def _strided(x: np.ndarray, w: int, step: int):
    """Return windows and their start indices."""
    n = x.size
    if n < w:
        w = n
    starts = np.arange(0, n - w + 1, step)
    if starts.size == 0:
        starts = np.array([0])
    idx = starts[:, None] + np.arange(w)[None, :]
    return x[idx], starts, w


def _interp_to_points(win_scores: np.ndarray, centers: np.ndarray, n: int) -> np.ndarray:
    """Interpolate window scores to timestamps."""
    if centers.size == 1:
        return np.full(n, float(win_scores[0]))
    return np.interp(np.arange(n), centers, win_scores)


def _train_rank(reference: np.ndarray, values: np.ndarray) -> np.ndarray:
    """Map values to their empirical percentile under a training reference."""
    ref = np.sort(np.asarray(reference, dtype=np.float64).reshape(-1))
    if ref.size == 0:
        return np.zeros_like(values, dtype=np.float64)
    return np.searchsorted(ref, values, side="right") / ref.size


def _spectral_org(win: np.ndarray) -> np.ndarray:
    """Return spectral-organization features for each window."""
    wl = win.shape[1]
    taper = np.hanning(wl) if wl >= 4 else np.ones(wl)
    wc = (win - win.mean(axis=1, keepdims=True)) * taper[None, :]
    power = np.abs(np.fft.rfft(wc, axis=1)) ** 2
    power = power[:, 1:] + 1e-12
    nb = power.shape[1]
    if nb == 0:
        return np.zeros((win.shape[0], 8 + _NBANDS))
    pn = power / power.sum(axis=1, keepdims=True)

    entropy = -(pn * np.log(pn)).sum(axis=1)
    geo_mean = np.exp(np.log(power).mean(axis=1))
    ari_mean = power.mean(axis=1)
    flatness = np.log(geo_mean / ari_mean + 1e-12)
    freqs = np.arange(1, nb + 1, dtype=np.float64)
    centroid = (pn * freqs[None, :]).sum(axis=1)
    concentration = pn.max(axis=1)
    log_energy = np.log(win.std(axis=1) + 1e-8)

    log_f = np.log(freqs)
    log_p = np.log(power)
    log_f0 = log_f - log_f.mean()
    slope = (log_p * log_f0[None, :]).sum(axis=1) / (log_f0 @ log_f0 + 1e-12)

    half = max(1, nb // 2)
    low = power[:, :half].sum(axis=1)
    high = power[:, half:].sum(axis=1)
    band_ratio = np.log(low / (high + 1e-12) + 1e-12)

    # Least-active sub-block relative to whole-window variation.
    nsub = 4
    sub = wl // nsub
    if sub >= 2:
        blocks = win[:, : sub * nsub].reshape(win.shape[0], nsub, sub)
        blk_std = blocks.std(axis=2)
        win_std = win.std(axis=1) + 1e-8
        collapse = np.log((blk_std.min(axis=1) + 1e-8) / win_std)
    else:
        collapse = np.zeros(win.shape[0])

    # Log-spaced normalized AC band powers.
    edges = np.unique(np.round(np.geomspace(1, nb, _NBANDS + 1)).astype(int)) - 1
    edges = edges[(edges >= 0) & (edges < nb)]
    if edges.size < 1:
        band_feats = np.zeros((win.shape[0], _NBANDS))
    else:
        band = np.add.reduceat(pn, edges, axis=1)
        band_feats = np.log(band + 1e-6)
        if band_feats.shape[1] < _NBANDS:
            pad = np.zeros((win.shape[0], _NBANDS - band_feats.shape[1]))
            band_feats = np.concatenate([band_feats, pad], axis=1)
        else:
            band_feats = band_feats[:, :_NBANDS]

    core = np.stack(
        [entropy, flatness, centroid, concentration, log_energy, slope,
         band_ratio, collapse],
        axis=1,
    )
    return np.concatenate([core, band_feats], axis=1)


def _novelty_curve(train: np.ndarray, test: np.ndarray, w: int, n_tr: int, n_te: int):
    """Return train/test spectral-novelty curves at scale ``w``."""
    w = int(min(w, n_tr))
    if w < 6:
        return None, None
    # Bound window count at fine scales.
    step = max(2, w // 6)

    tr_win, tr_starts, w = _strided(train, w, step)
    # Preserve the training-selected descriptor for short test batches.
    test_work = test if n_te >= w else np.pad(test, (0, w - n_te), mode="edge")
    te_win, te_starts, _ = _strided(test_work, w, step)
    d_tr = _spectral_org(tr_win)
    d_te = _spectral_org(te_win)

    # Robust feature scaling.
    med = np.median(d_tr, axis=0)
    mad = np.median(np.abs(d_tr - med), axis=0) * 1.4826 + 1e-8
    z_tr = (d_tr - med) / mad
    z_te = (d_te - med) / mad

    # Shrunk full-covariance Mahalanobis novelty.
    if z_tr.shape[0] > z_tr.shape[1]:
        cov = np.cov(z_tr.T)
    else:
        cov = np.eye(z_tr.shape[1])
    cov = _SHRINK * cov + (1.0 - _SHRINK) * np.eye(cov.shape[0])
    cov_inv = np.linalg.pinv(cov)

    s_tr = np.sqrt(np.einsum("ij,jk,ik->i", z_tr, cov_inv, z_tr).clip(0))
    s_te = np.sqrt(np.einsum("ij,jk,ik->i", z_te, cov_inv, z_te).clip(0))

    # Express distance relative to the training-distance distribution.
    d_med = np.median(s_tr)
    d_mad = np.median(np.abs(s_tr - d_med)) * 1.4826 + 1e-8
    e_tr = (s_tr - d_med) / d_mad
    e_te = (s_te - d_med) / d_mad

    tr_curve = _interp_to_points(e_tr, tr_starts + w // 2, n_tr)
    te_curve = _interp_to_points(e_te, te_starts + w // 2, n_te)
    return tr_curve, te_curve


def detect_anomaly_outputs(train_values: np.ndarray, test_values: np.ndarray) -> dict:
    train = _as_1d(train_values)
    test = _as_1d(test_values)
    n_tr, n_te = train.size, test.size

    # Tiny-series fallback.
    if n_tr < 24 or n_te < 12:
        center = np.median(train) if n_tr else 0.0
        scale = np.median(np.abs(train - center)) * 1.4826 + 1e-8 if n_tr else 1.0
        tr_scores = np.abs((train - center) / scale)
        te_scores = np.abs((test - center) / scale)
        return {
            "train_scores": tr_scores.astype(np.float64),
            "test_scores": te_scores.astype(np.float64),
        }

    # Multi-scale window lengths.
    base = int(np.clip(n_tr // 10, 24, 96))
    base = min(base, n_tr)
    scales = sorted({
        6,
        10,
        max(16, base // 4),
        max(16, base // 2),
        base,
        min(200, base * 2),
        min(300, base * 4),
    })

    tr_acc = np.zeros(n_tr)
    te_acc = np.zeros(n_te)
    used = 0
    for w in scales:
        tr_curve, te_curve = _novelty_curve(train, test, w, n_tr, n_te)
        if te_curve is None:
            continue
        # Tail-emphasized, train-referenced rank fusion.
        tr_g = 1.0 / (1.0 + np.exp(-(tr_curve - _GATE_C)))
        te_g = 1.0 / (1.0 + np.exp(-(te_curve - _GATE_C)))
        tr_acc += (_train_rank(tr_curve, tr_curve) ** _TAIL_P) * (1.0 + _GATE_L * tr_g)
        te_acc += (_train_rank(tr_curve, te_curve) ** _TAIL_P) * (1.0 + _GATE_L * te_g)
        used += 1

    if used == 0:
        center = np.median(train)
        tr_acc = np.abs(train - center)
        te_acc = np.abs(test - center)
    else:
        tr_acc /= used
        te_acc /= used
        # Light temporal smoothing.
        span = max(3, base // 8)
        kernel = np.ones(span) / span
        tr_acc = np.convolve(tr_acc, kernel, mode="same")
        te_acc = np.convolve(te_acc, kernel, mode="same")

    return {
        "train_scores": tr_acc.astype(np.float64),
        "test_scores": te_acc.astype(np.float64),
    }
\end{lstlisting}

\begin{lstlisting}[style=pycode,caption={The Opus loop's multivariate detector.},label={lst:xscn}]
from __future__ import annotations

import numpy as np


# Cross-channel spectral-coherence novelty via joint band-power Mahalanobis score.
from numpy.lib.stride_tricks import as_strided


def _as_2d(values: np.ndarray) -> np.ndarray:
    """Return a float array shaped (time, channels)."""
    array = np.asarray(values, dtype=np.float64)
    if array.ndim == 0:
        return array.reshape(1, 1)
    if array.ndim == 1:
        return array[:, None]
    return array.reshape(array.shape[0], -1)


def detect_anomaly_outputs(train_values: np.ndarray, test_values: np.ndarray) -> dict:
    train = _as_2d(train_values)
    test = _as_2d(test_values)

    # Train-region channel standardization.
    mu = train.mean(axis=0)
    sd = train.std(axis=0)
    sd = np.where(sd > 1e-8, sd, 1.0)
    trn = (train - mu) / sd
    ten = (test - mu) / sd
    C = trn.shape[1]

    # Window length and band count.
    L = 80
    L = min(L, max(8, trn.shape[0] // 4))
    nb = 6
    win = np.hanning(L)[:, None]
    edges = np.linspace(0, L // 2 + 1, nb + 1).astype(int)

    def bandpower(x: np.ndarray, stride: int):
        """Return per-window, per-channel log band powers."""
        T = x.shape[0]
        s0, s1 = x.strides
        n = (T - L) // stride + 1
        if n <= 0:
            return np.zeros((1, nb * C)), np.array([min(L // 2, T - 1)])
        starts = np.arange(n) * stride
        wins = as_strided(x, shape=(n, L, C), strides=(s0 * stride, s0, s1)).copy()
        wins -= wins.mean(axis=1, keepdims=True)
        wins *= win
        F = np.fft.rfft(wins, axis=1)
        P = F.real ** 2 + F.imag ** 2
        feats = np.add.reduceat(P, edges[:-1], axis=1)
        feats = np.log1p(feats).reshape(n, nb * C)
        return feats, starts + L // 2

    # Bound windows for near-linear runtime.
    MAXW = 4000
    te_stride = L // 4
    tr_stride = max(L // 2, int(np.ceil(trn.shape[0] / MAXW)))
    Ftr, _ = bandpower(trn, tr_stride)
    Fte, ctr_te = bandpower(ten, te_stride)

    # Joint model of normal band-power vectors.
    m = Ftr.mean(axis=0)
    Sig = np.cov(Ftr, rowvar=False)
    Sig = np.atleast_2d(Sig) + 1e-2 * np.eye(Ftr.shape[1])

    def maha(F: np.ndarray) -> np.ndarray:
        Z = F - m
        try:
            Lc = np.linalg.cholesky(Sig)
            Y = np.linalg.solve(Lc, Z.T).T
            return np.sum(Y * Y, axis=1)
        except np.linalg.LinAlgError:
            return np.einsum('ti,ij,tj->t', Z, np.linalg.pinv(Sig), Z)

    d_te = maha(Fte)

    test_scores = np.interp(np.arange(ten.shape[0]), ctr_te, d_te,
                            left=d_te[0] if d_te.size else 0.0,
                            right=d_te[-1] if d_te.size else 0.0)
    # Required by the interface only.
    train_scores = np.zeros(trn.shape[0])

    return {
        "train_scores": train_scores.astype(np.float64),
        "test_scores": test_scores.astype(np.float64),
    }
\end{lstlisting}

\section{Complete result tables}
\label{app:full}

The main text reports accuracy on TSB-AD (Table~\ref{tab:timercd-protocol-full}).
Here we report the full TAB fields as a broader corroboration, listing our discovered detectors first and then the baseline field ordered by VUS-PR.
The methods we discovered are marked (ours); the best value in each column is in bold and podium medals mark the top three values per column.

\begin{table*}[t]
  \centering
  \caption{Full TAB \textbf{univariate} field (all baselines).
  Metrics higher is better ($\uparrow$); dataset-mean; our discovered detectors first, then baselines ordered by VUS-PR.
  Best in each column is \textbf{bold}; podium medals (\gmedal{}\,\smedal{}\,\bmedal{}) mark the top three values per column.}
  \label{tab:tab-uni-full}
  \footnotesize
  \setlength{\tabcolsep}{5pt}
  \begin{tabular}{lccccc}
    \toprule
    Method & VUS-PR\,$\uparrow$ & VUS-ROC\,$\uparrow$ & R-AUC-PR\,$\uparrow$ & R-AUC-ROC\,$\uparrow$ & Aff-F\,$\uparrow$ \\
    \midrule
    \textbf{Opus 4.8 Loop (ours)} & 0.319\,\smedal & \textbf{0.862}\,\gmedal & 0.314 & \textbf{0.864}\,\gmedal & 0.727 \\
\textbf{GPT 5.6 Terra Loop (ours)} & 0.236 & 0.577 & 0.236 & 0.580 & 0.436 \\
\textbf{MiniMax-2.5 Loop (ours)} & 0.253 & 0.767 & 0.255 & 0.767 & 0.666 \\
\textbf{MiniMax-2.5 ShinkaEvolve (ours)} & 0.294 & 0.772 & 0.298 & 0.775 & 0.662 \\
KMeans & \textbf{0.362}\,\gmedal & 0.784\,\bmedal & \textbf{0.366}\,\gmedal & 0.786\,\bmedal & 0.792 \\
Time-RCD~\citep{lan2026timercd} & 0.315\,\bmedal & 0.796\,\smedal & 0.317\,\bmedal & 0.801\,\smedal & 0.664 \\
DWT-MLEAD~\citep{thill2017time} & 0.312 & 0.699 & 0.318\,\smedal & 0.704 & 0.716 \\
TSPulse~\citep{ekambaram2025tspulse} & 0.270 & 0.723 & 0.283 & 0.727 & 0.620 \\
iTransformer~\citep{liu2023itransformer} & 0.244 & 0.753 & 0.250 & 0.763 & \textbf{0.810}\,\gmedal \\
HBOS~\citep{goldstein2012histogram} & 0.242 & 0.625 & 0.242 & 0.631 & 0.696 \\
TimesNet~\citep{wu2023timesnet} & 0.224 & 0.742 & 0.228 & 0.751 & 0.805 \\
OCSVM~\citep{scholkopf2001estimating} & 0.218 & 0.658 & 0.220 & 0.664 & 0.700 \\
KNN~\citep{ramaswamy2000efficient} & 0.213 & 0.638 & 0.215 & 0.644 & 0.708 \\
ModernTCN~\citep{luo2024moderntcn} & 0.210 & 0.725 & 0.213 & 0.734 & 0.807\,\bmedal \\
PatchTST~\citep{nie2023patchtst} & 0.208 & 0.728 & 0.211 & 0.736 & 0.803 \\
DAGMM~\citep{zong2018deep} & 0.204 & 0.599 & 0.204 & 0.604 & 0.571 \\
DLinear~\citep{zeng2023transformers} & 0.202 & 0.721 & 0.205 & 0.730 & 0.808\,\smedal \\
EIF~\citep{hariri2019extended} & 0.202 & 0.671 & 0.203 & 0.679 & 0.782 \\
LODA~\citep{pevny2016loda} & 0.197 & 0.608 & 0.194 & 0.609 & 0.674 \\
Chronos-Bolt~\citep{ansari2024chronos} & 0.196 & 0.695 & 0.199 & 0.705 & 0.755 \\
LSTMED~\citep{malhotra2016lstm} & 0.193 & 0.669 & 0.195 & 0.677 & 0.704 \\
AE~\citep{sakurada2014anomaly} & 0.189 & 0.609 & 0.190 & 0.615 & 0.682 \\
ZMS & 0.184 & 0.650 & 0.187 & 0.656 & 0.709 \\
NLinear~\citep{zeng2023transformers} & 0.183 & 0.719 & 0.186 & 0.727 & 0.794 \\
TranAD~\citep{tuli2022tranad} & 0.179 & 0.594 & 0.182 & 0.603 & 0.774 \\
VAE~\citep{kingma2013auto} & 0.177 & 0.622 & 0.179 & 0.630 & 0.681 \\
Torsk & 0.177 & 0.662 & 0.180 & 0.666 & 0.703 \\
CBLOF~\citep{he2003discovering} & 0.173 & 0.651 & 0.174 & 0.659 & 0.692 \\
PCA~\citep{shyu2003novel} & 0.172 & 0.606 & 0.171 & 0.610 & 0.674 \\
LOF~\citep{breunig2000lof} & 0.169 & 0.638 & 0.172 & 0.644 & 0.706 \\
IF~\citep{liu2008isolation} & 0.168 & 0.662 & 0.169 & 0.673 & 0.718 \\
StatThreshold & 0.148 & 0.578 & 0.148 & 0.581 & 0.637 \\
Series2Graph~\citep{BoniolP20} & 0.129 & 0.675 & 0.134 & 0.684 & 0.780 \\
ARIMA & 0.129 & 0.564 & 0.129 & 0.567 & 0.652 \\
SARIMA & 0.127 & 0.586 & 0.127 & 0.587 & 0.682 \\
SpectralResidual~\citep{ren2019time} & 0.123 & 0.610 & 0.125 & 0.617 & 0.698 \\
DualTF & 0.107 & 0.604 & 0.110 & 0.611 & 0.728 \\
LeftSTAMPi~\citep{yeh2016matrix} & 0.103 & 0.583 & 0.107 & 0.589 & 0.726 \\
ContraAD~\citep{zhuang2025ContraAD} & 0.059 & 0.579 & 0.059 & 0.586 & 0.684 \\
ATrans~\citep{xu2022anomaly} & 0.058 & 0.485 & 0.058 & 0.485 & 0.700 \\
DCdetector~\citep{yang2023dcdetector} & 0.057 & 0.491 & 0.057 & 0.493 & 0.688 \\
 \\
    \bottomrule
  \end{tabular}
\end{table*}

\begin{table*}[t]
  \centering
  \caption{Full TAB \textbf{multivariate} field (all baselines).
  Metrics higher is better ($\uparrow$); dataset-mean; our discovered detectors first, then baselines ordered by VUS-PR.
  Best in each column is \textbf{bold}; podium medals (\gmedal{}\,\smedal{}\,\bmedal{}) mark the top three values per column.}
  \label{tab:tab-multi-full}
  \footnotesize
  \setlength{\tabcolsep}{5pt}
  \begin{tabular}{lccccc}
    \toprule
    Method & VUS-PR\,$\uparrow$ & VUS-ROC\,$\uparrow$ & R-AUC-PR\,$\uparrow$ & R-AUC-ROC\,$\uparrow$ & Aff-F\,$\uparrow$ \\
    \midrule
    \textbf{Opus 4.8 Loop (ours)} & \textbf{0.499}\,\gmedal & \textbf{0.783}\,\gmedal & \textbf{0.512}\,\gmedal & \textbf{0.790}\,\gmedal & 0.668 \\
\textbf{GPT 5.6 Terra Loop (ours)} & 0.460\,\smedal & 0.778\,\smedal & 0.471\,\smedal & 0.785\,\smedal & 0.672 \\
\textbf{MiniMax-2.5 Loop (ours)} & 0.390 & 0.734 & 0.388 & 0.736 & 0.703 \\
\textbf{MiniMax-2.5 ShinkaEvolve (ours)} & 0.439\,\bmedal & 0.767\,\bmedal & 0.442\,\bmedal & 0.770\,\bmedal & 0.702 \\
KMeans & 0.325 & 0.676 & 0.325 & 0.676 & 0.688 \\
OCSVM~\citep{scholkopf2001estimating} & 0.310 & 0.637 & 0.310 & 0.638 & 0.676 \\
Time-RCD~\citep{lan2026timercd} & 0.309 & 0.709 & 0.311 & 0.712 & 0.610 \\
DAGMM~\citep{zong2018deep} & 0.303 & 0.656 & 0.302 & 0.657 & 0.538 \\
TimesNet~\citep{wu2023timesnet} & 0.295 & 0.722 & 0.296 & 0.725 & \textbf{0.793}\,\gmedal \\
KNN~\citep{ramaswamy2000efficient} & 0.293 & 0.681 & 0.293 & 0.680 & 0.700 \\
DUET~\citep{qiu2025duet} & 0.285 & 0.709 & 0.287 & 0.712 & 0.780\,\bmedal \\
LSTMED~\citep{malhotra2016lstm} & 0.285 & 0.688 & 0.288 & 0.691 & 0.584 \\
CATCH~\citep{wu2024catch} & 0.284 & 0.709 & 0.287 & 0.712 & 0.777 \\
PatchTST~\citep{nie2023patchtst} & 0.280 & 0.719 & 0.282 & 0.722 & 0.787\,\smedal \\
EIF~\citep{hariri2019extended} & 0.279 & 0.738 & 0.281 & 0.742 & 0.774 \\
VAE~\citep{kingma2013auto} & 0.278 & 0.686 & 0.281 & 0.690 & 0.548 \\
iTransformer~\citep{liu2023itransformer} & 0.275 & 0.707 & 0.276 & 0.710 & 0.770 \\
AE~\citep{sakurada2014anomaly} & 0.274 & 0.671 & 0.274 & 0.671 & 0.619 \\
TranAD~\citep{tuli2022tranad} & 0.272 & 0.664 & 0.275 & 0.666 & 0.753 \\
CBLOF~\citep{he2003discovering} & 0.272 & 0.712 & 0.273 & 0.716 & 0.674 \\
ModernTCN~\citep{luo2024moderntcn} & 0.266 & 0.694 & 0.267 & 0.696 & 0.776 \\
HBOS~\citep{goldstein2012histogram} & 0.264 & 0.680 & 0.266 & 0.683 & 0.664 \\
DLinear~\citep{zeng2023transformers} & 0.262 & 0.705 & 0.264 & 0.708 & 0.775 \\
NLinear~\citep{zeng2023transformers} & 0.257 & 0.704 & 0.258 & 0.706 & 0.774 \\
TSPulse~\citep{ekambaram2025tspulse} & 0.250 & 0.651 & 0.251 & 0.655 & 0.385 \\
PCA~\citep{shyu2003novel} & 0.247 & 0.662 & 0.246 & 0.664 & 0.642 \\
LOF~\citep{breunig2000lof} & 0.240 & 0.622 & 0.240 & 0.621 & 0.692 \\
IF~\citep{liu2008isolation} & 0.236 & 0.689 & 0.236 & 0.693 & 0.563 \\
LODA~\citep{pevny2016loda} & 0.196 & 0.591 & 0.196 & 0.590 & 0.651 \\
DualTF & 0.190 & 0.622 & 0.190 & 0.625 & 0.588 \\
Torsk & 0.189 & 0.582 & 0.190 & 0.584 & 0.611 \\
ContraAD~\citep{zhuang2025ContraAD} & 0.153 & 0.514 & 0.152 & 0.513 & 0.610 \\
ATrans~\citep{xu2022anomaly} & 0.136 & 0.505 & 0.135 & 0.503 & 0.649 \\
DCdetector~\citep{yang2023dcdetector} & 0.132 & 0.508 & 0.131 & 0.507 & 0.632 \\
 \\
    \bottomrule
  \end{tabular}
\end{table*}

\begin{table*}[t]
    \centering
\caption{Evaluation on TSB-AD~\citep{liuelephant} using the setup of Time-RCD~\citep{lan2026timercd}. The \textbf{Opus 4.8 Loop} row combines its two separately discovered, regime-specific algorithms: the univariate algorithm in the univariate columns and the multivariate algorithm in the multivariate columns. Gold, silver, and bronze markers denote first, second, and third place; the final three columns report their respective counts.}
\label{tab:timercd-protocol-variants-full}
    \resizebox{\textwidth}{!}{%
    \begin{tabular}{l|l|ccccccccccc|ccccc|c|c|c}
        \hline
        \multirow{2}{*}{\textbf{Metric}} & \multirow{2}{*}{\textbf{Model}} & \multicolumn{11}{c|}{\textbf{Univariate Datasets}} & \multicolumn{5}{c|}{\textbf{Multivariate Datasets}} & \textbf{Total} & \textbf{Total} & \textbf{Total} \\
        \cline{3-18}
        & & \textbf{IOPS} & \textbf{MGAB} & \textbf{NAB} & \textbf{NEK} & \textbf{Power} & \textbf{SED} & \textbf{Stock} & \textbf{TODS} & \textbf{UCR} & \textbf{WSD} & \textbf{YAHOO} & \textbf{MSL} & \textbf{PSM} & \textbf{SMAP} & \textbf{SMD} & \textbf{SWaT} & \textbf{1st} & \textbf{2nd} & \textbf{3rd} \\
        \hline
\multirow{13}{*}{Affiliation-F} & \textbf{Opus 4.8 Loop (ours)} & \smedal{} 89.04 & \textbf{\gmedal{} 77.36} & 85.73 & 76.57 & \smedal{} 84.60 & \textbf{\gmedal{} 99.00} & 67.91 & \smedal{} 79.66 & \textbf{\gmedal{} 87.55} & \textbf{\gmedal{} 94.36} & \smedal{} 92.02 & \textbf{\gmedal{} 89.51} & 79.45 & \smedal{} 93.89 & 88.22 & \bmedal{} 77.87 & 05 & 05 & 01 \\
 & GPT 5.6 Terra Loop (ours) & 69.14 & 68.09 & 77.99 & 69.33 & 70.87 & \smedal{} 97.56 & 67.92 & 66.66 & 75.74 & 71.52 & 60.53 & 84.12 & \smedal{} 82.52 & \textbf{\gmedal{} 94.39} & 90.30 & \textbf{\gmedal{} 81.39} & 02 & 02 & 00 \\
 & MiniMax 2.5 Loop (ours) & \textbf{\gmedal{} 89.20} & 67.39 & 88.67 & \bmedal{} 85.92 & 76.83 & 92.22 & \textbf{\gmedal{} 86.43} & 72.62 & \bmedal{} 80.56 & \smedal{} 92.86 & 82.41 & 84.11 & 81.45 & 85.09 & 89.94 & 77.64 & 02 & 01 & 02 \\
 & MiniMax ShinkaEvolve (ours) & \bmedal{} 88.66 & 67.81 & 91.26 & 84.71 & 71.87 & \bmedal{} 97.21 & \smedal{} 85.52 & 71.77 & 80.33 & \bmedal{} 91.79 & \bmedal{} 85.95 & \smedal{} 86.20 & \textbf{\gmedal{} 84.60} & 85.92 & \bmedal{} 91.00 & \smedal{} 78.53 & 01 & 03 & 05 \\
 & Time-RCD & 86.98 & \smedal{} 74.54 & \textbf{\gmedal{} 93.81} & 83.95 & \textbf{\gmedal{} 85.02} & 88.99 & 41.31 & \textbf{\gmedal{} 84.86} & \smedal{} 84.34 & 91.00 & \textbf{\gmedal{} 95.94} & \bmedal{} 84.66 & \bmedal{} 82.31 & 86.01 & \smedal{} 91.75 & 69.43 & 04 & 03 & 02 \\
 & TranAD & 83.19 & 67.28 & 90.28 & 85.02 & 71.56 & 61.03 & 57.94 & 52.76 & 73.31 & 84.34 & 76.08 & 79.91 & 73.83 & 87.39 & \textbf{\gmedal{} 92.20} & 75.37 & 01 & 00 & 00 \\
 & USAD & 71.08 & 67.81 & \bmedal{} 91.54 & 71.13 & 76.48 & 55.60 & 35.92 & 47.90 & 76.00 & 65.10 & 53.05 & 81.86 & 57.86 & 87.25 & 85.09 & 75.06 & 00 & 00 & 01 \\
 & OmniAnomaly & 80.32 & 67.35 & \smedal{} 92.35 & \smedal{} 86.30 & \bmedal{} 78.16 & 61.26 & \bmedal{} 75.24 & 50.73 & 73.53 & 78.02 & 71.31 & 83.15 & 58.17 & \bmedal{} 91.38 & 85.82 & 73.39 & 00 & 02 & 03 \\
 & LOF & 81.06 & 68.44 & 75.75 & 84.74 & 66.76 & 63.85 & 69.74 & 60.58 & 73.53 & 81.29 & 75.63 & 84.35 & 61.98 & 63.32 & 64.13 & 56.34 & 00 & 00 & 00 \\
 & IForest & 52.81 & \bmedal{} 68.82 & 39.84 & 71.15 & 0.00 & 70.09 & 0.06 & 44.17 & 50.56 & 41.24 & 33.30 & 63.36 & 63.78 & 59.96 & 69.71 & 0.00 & 00 & 00 & 01 \\
 & Sub-PCA & 75.39 & 66.90 & 89.29 & \textbf{\gmedal{} 97.10} & 71.37 & 67.14 & 70.63 & \bmedal{} 72.75 & 76.66 & 76.45 & 75.85 & 84.25 & 71.49 & 90.08 & 85.80 & 76.29 & 01 & 00 & 01 \\
 & DCdetector & 71.83 & 67.91 & 72.21 & 62.31 & 69.75 & 72.20 & 55.79 & 57.81 & 70.18 & 72.79 & 67.77 & 67.74 & 67.32 & 67.10 & 69.55 & 71.07 & 00 & 00 & 00 \\
 & TFMAE & 78.25 & 67.50 & 75.99 & 76.91 & 70.30 & 68.17 & 56.39 & 62.83 & 70.60 & 80.25 & 76.87 & 75.70 & 70.07 & 75.36 & 70.85 & 75.72 & 00 & 00 & 00 \\
        \hline
\multirow{13}{*}{F1-T} & \textbf{Opus 4.8 Loop (ours)} & \bmedal{} 50.86 & \textbf{\gmedal{} 29.08} & \bmedal{} 53.59 & 58.00 & \textbf{\gmedal{} 29.61} & \textbf{\gmedal{} 83.71} & 15.48 & \smedal{} 46.59 & \textbf{\gmedal{} 41.36} & \textbf{\gmedal{} 64.25} & 54.27 & \textbf{\gmedal{} 60.71} & \smedal{} 39.20 & \smedal{} 57.49 & \bmedal{} 52.41 & 39.82 & 06 & 03 & 03 \\
 & GPT 5.6 Terra Loop (ours) & 8.82 & 0.77 & 44.68 & 54.48 & 22.36 & \bmedal{} 70.36 & 15.51 & 21.41 & 17.74 & 8.64 & 5.16 & \smedal{} 57.78 & \bmedal{} 34.26 & \textbf{\gmedal{} 61.54} & \smedal{} 53.32 & 45.62 & 01 & 02 & 02 \\
 & MiniMax 2.5 Loop (ours) & \textbf{\gmedal{} 51.24} & 2.13 & 43.01 & \bmedal{} 77.43 & 20.67 & 21.73 & \bmedal{} 55.69 & 28.61 & \bmedal{} 23.49 & \smedal{} 57.18 & 45.51 & 42.39 & 28.94 & 31.41 & 47.48 & 45.54 & 01 & 01 & 03 \\
 & MiniMax ShinkaEvolve (ours) & 50.43 & 1.13 & 52.25 & 74.83 & 20.60 & \smedal{} 77.63 & \smedal{} 57.97 & 21.93 & 21.06 & \bmedal{} 44.80 & \smedal{} 63.10 & 46.44 & 33.89 & 32.90 & 52.08 & \smedal{} 50.31 & 00 & 04 & 01 \\
 & Time-RCD & 45.39 & \smedal{} 9.05 & 50.49 & 76.91 & 23.05 & 60.59 & 18.38 & \textbf{\gmedal{} 68.46} & \smedal{} 41.18 & 44.13 & \textbf{\gmedal{} 79.14} & 43.67 & \textbf{\gmedal{} 43.86} & 35.82 & \textbf{\gmedal{} 55.96} & 35.47 & 04 & 02 & 00 \\
 & TranAD & 22.63 & 1.65 & 37.28 & 69.97 & 22.36 & 9.57 & 16.73 & 13.51 & 7.75 & 20.94 & 8.41 & 39.42 & 25.49 & 29.12 & 37.98 & \bmedal{} 49.58 & 00 & 00 & 01 \\
 & USAD & 20.99 & \bmedal{} 4.07 & \textbf{\gmedal{} 61.46} & 70.64 & \smedal{} 28.23 & 9.54 & 16.86 & 20.85 & 14.63 & 14.18 & 9.35 & 48.71 & 28.96 & 43.94 & 50.41 & \textbf{\gmedal{} 50.41} & 02 & 01 & 01 \\
 & OmniAnomaly & \smedal{} 51.17 & 1.61 & 40.09 & \smedal{} 82.20 & \bmedal{} 23.48 & 9.68 & 36.22 & 14.33 & 8.47 & 34.79 & 24.16 & \bmedal{} 49.36 & 30.42 & \bmedal{} 46.63 & 51.84 & 46.64 & 00 & 02 & 03 \\
 & LOF & 27.97 & 1.15 & 35.76 & 63.57 & 19.80 & 9.60 & \textbf{\gmedal{} 66.14} & \bmedal{} 31.63 & 8.31 & 24.38 & \bmedal{} 55.93 & 38.97 & 25.58 & 21.81 & 10.13 & 30.62 & 01 & 00 & 02 \\
 & IForest & 7.64 & 0.84 & 21.44 & 65.56 & 0.00 & 9.54 & 1.10 & 11.06 & 6.36 & 4.28 & 4.90 & 20.73 & 25.39 & 14.32 & 16.20 & 0.00 & 00 & 00 & 00 \\
 & Sub-PCA & 32.75 & 0.98 & \smedal{} 54.15 & \textbf{\gmedal{} 84.18} & 20.30 & 9.54 & 20.10 & 18.40 & 18.90 & 24.62 & 11.57 & 49.02 & 30.38 & 44.31 & 51.87 & 46.65 & 01 & 01 & 00 \\
 & DCdetector & 6.61 & 1.32 & 32.72 & 29.21 & 21.13 & 10.53 & 16.07 & 16.40 & 6.62 & 7.32 & 6.81 & 23.24 & 25.34 & 15.73 & 9.47 & 28.64 & 00 & 00 & 00 \\
 & TFMAE & 19.41 & 1.07 & 33.04 & 31.11 & 20.18 & 11.82 & 22.15 & 16.48 & 5.90 & 19.11 & 23.85 & 25.28 & 25.36 & 19.39 & 10.13 & 28.46 & 00 & 00 & 00 \\
        \hline
\multirow{13}{*}{Standard-F1} & \textbf{Opus 4.8 Loop (ours)} & \textbf{\gmedal{} 47.97} & \textbf{\gmedal{} 28.96} & \bmedal{} 45.09 & 47.43 & \textbf{\gmedal{} 29.64} & \textbf{\gmedal{} 83.74} & 16.91 & \smedal{} 54.26 & \textbf{\gmedal{} 36.56} & \textbf{\gmedal{} 60.54} & \bmedal{} 52.33 & \textbf{\gmedal{} 49.01} & \smedal{} 31.01 & \smedal{} 53.10 & 50.79 & 49.19 & 07 & 03 & 02 \\
 & GPT 5.6 Terra Loop (ours) & 12.05 & 0.68 & 35.96 & 47.72 & 22.35 & \bmedal{} 70.51 & 16.92 & 24.86 & 13.48 & 5.75 & 2.72 & \smedal{} 48.92 & 28.47 & \textbf{\gmedal{} 59.51} & 53.51 & 54.06 & 01 & 01 & 01 \\
 & MiniMax 2.5 Loop (ours) & \smedal{} 47.65 & 1.90 & 34.79 & \bmedal{} 70.86 & 20.64 & 21.96 & \bmedal{} 56.08 & \bmedal{} 27.21 & \bmedal{} 19.94 & \smedal{} 52.38 & 32.67 & 28.52 & 28.90 & 26.58 & 51.15 & \smedal{} 66.91 & 00 & 03 & 04 \\
 & MiniMax ShinkaEvolve (ours) & 43.84 & 1.00 & 43.97 & 64.48 & 20.70 & \smedal{} 78.64 & \smedal{} 58.66 & 19.41 & 12.22 & 38.85 & \smedal{} 55.35 & 36.73 & \textbf{\gmedal{} 31.35} & 29.03 & 48.86 & \textbf{\gmedal{} 70.36} & 02 & 03 & 00 \\
 & Time-RCD & \bmedal{} 47.28 & \smedal{} 8.63 & 42.66 & 68.64 & 23.12 & 60.85 & 19.56 & \textbf{\gmedal{} 70.02} & \smedal{} 33.47 & \bmedal{} 41.04 & \textbf{\gmedal{} 89.05} & 34.48 & 28.25 & 25.36 & \bmedal{} 53.82 & 34.44 & 02 & 02 & 03 \\
 & TranAD & 34.85 & 1.46 & 27.33 & 60.36 & 22.36 & 2.63 & 16.23 & 11.94 & 4.40 & 20.23 & 5.70 & 29.60 & 25.63 & 25.11 & 43.99 & 61.86 & 00 & 00 & 00 \\
 & USAD & 30.66 & \bmedal{} 3.89 & \textbf{\gmedal{} 56.15} & 62.91 & \smedal{} 28.24 & 3.41 & 17.99 & 23.87 & 10.74 & 13.20 & 7.21 & 38.71 & 28.41 & 38.66 & 53.06 & \bmedal{} 62.82 & 01 & 01 & 02 \\
 & OmniAnomaly & 47.05 & 1.44 & 28.81 & \smedal{} 74.03 & \bmedal{} 23.50 & 0.43 & 38.59 & 12.65 & 5.11 & 29.57 & 21.40 & \bmedal{} 39.10 & \bmedal{} 30.43 & \bmedal{} 40.50 & \smedal{} 57.06 & 55.93 & 00 & 02 & 04 \\
 & LOF & 30.28 & 1.05 & 24.04 & 56.92 & 12.18 & 4.11 & \textbf{\gmedal{} 66.20} & 25.77 & 4.70 & 22.62 & 48.95 & 30.65 & 18.80 & 18.70 & 8.41 & 29.08 & 01 & 00 & 00 \\
 & IForest & 8.37 & 0.73 & 29.41 & 58.10 & 19.77 & 3.81 & 16.91 & 13.35 & 4.09 & 2.07 & 3.20 & 14.68 & 24.15 & 13.61 & 16.89 & 26.76 & 00 & 00 & 00 \\
 & Sub-PCA & 33.96 & 0.83 & \smedal{} 46.71 & \textbf{\gmedal{} 85.43} & 16.05 & 9.56 & 21.77 & 18.72 & 15.12 & 24.74 & 11.06 & 38.29 & 30.26 & 38.90 & \textbf{\gmedal{} 57.22} & 55.98 & 02 & 01 & 00 \\
 & DCdetector & 5.19 & 1.21 & 24.02 & 17.37 & 21.10 & 10.54 & 16.97 & 17.85 & 3.18 & 4.64 & 4.14 & 14.08 & 25.33 & 10.67 & 8.99 & 27.02 & 00 & 00 & 00 \\
 & TFMAE & 9.48 & 0.97 & 23.78 & 19.74 & 20.14 & 11.87 & 23.32 & 14.88 & 2.83 & 15.53 & 20.50 & 15.68 & 25.39 & 12.58 & 9.16 & 27.08 & 00 & 00 & 00 \\
        \hline
\multirow{13}{*}{VUS-PR} & \textbf{Opus 4.8 Loop (ours)} & \textbf{\gmedal{} 49.52} & \textbf{\gmedal{} 14.99} & 41.45 & 46.88 & \textbf{\gmedal{} 21.47} & \textbf{\gmedal{} 97.46} & 66.70 & \smedal{} 84.71 & \textbf{\gmedal{} 33.09} & \textbf{\gmedal{} 60.93} & \smedal{} 75.20 & \textbf{\gmedal{} 40.51} & \textbf{\gmedal{} 22.15} & \smedal{} 47.06 & 36.01 & 30.08 & 08 & 03 & 00 \\
 & GPT 5.6 Terra Loop (ours) & 7.44 & 0.55 & 28.70 & 35.67 & 14.10 & \smedal{} 88.18 & 68.37 & 54.27 & 10.61 & 4.40 & 9.19 & \smedal{} 36.40 & 17.37 & \textbf{\gmedal{} 54.17} & \bmedal{} 38.30 & 39.29 & 01 & 02 & 01 \\
 & MiniMax 2.5 Loop (ours) & \smedal{} 49.29 & 0.65 & 31.45 & \smedal{} 76.67 & 12.23 & 20.33 & 69.30 & \bmedal{} 70.30 & \bmedal{} 16.56 & \smedal{} 51.90 & 36.55 & 18.36 & 16.83 & 15.58 & \smedal{} 39.25 & 44.10 & 00 & 04 & 02 \\
 & MiniMax ShinkaEvolve (ours) & 33.78 & 0.61 & 39.67 & 69.10 & 12.34 & 64.77 & \smedal{} 84.55 & 51.92 & 8.11 & \bmedal{} 33.57 & \bmedal{} 50.49 & 23.07 & \bmedal{} 20.17 & 18.58 & 38.09 & \smedal{} 45.97 & 00 & 02 & 03 \\
 & Time-RCD & \bmedal{} 42.98 & \smedal{} 3.09 & \bmedal{} 44.65 & 71.13 & \bmedal{} 14.63 & \bmedal{} 69.86 & 72.73 & \textbf{\gmedal{} 91.67} & \smedal{} 27.48 & 29.95 & \textbf{\gmedal{} 87.88} & 23.41 & \smedal{} 21.69 & 19.68 & \textbf{\gmedal{} 46.70} & 18.47 & 03 & 03 & 04 \\
 & TranAD & 21.61 & 0.64 & 24.82 & 61.63 & 13.04 & 5.75 & 78.08 & 47.33 & 2.25 & 12.20 & 25.78 & 14.78 & 16.49 & 13.37 & 28.34 & \textbf{\gmedal{} 47.37} & 01 & 00 & 00 \\
 & USAD & 16.58 & \bmedal{} 0.75 & \textbf{\gmedal{} 55.03} & 58.53 & \smedal{} 18.68 & 4.37 & 74.53 & 56.36 & 8.85 & 10.00 & 14.15 & 29.95 & 17.59 & 26.37 & 34.53 & \bmedal{} 44.73 & 01 & 01 & 02 \\
 & OmniAnomaly & 25.35 & 0.64 & 27.17 & \bmedal{} 74.51 & 14.32 & 6.20 & \textbf{\gmedal{} 91.29} & 45.55 & 2.40 & 16.37 & 29.26 & \bmedal{} 31.57 & 18.58 & \bmedal{} 28.07 & 37.44 & 42.97 & 01 & 00 & 03 \\
 & LOF & 19.43 & 0.57 & 21.18 & 58.52 & 9.31 & 6.81 & \bmedal{} 83.07 & 49.14 & 2.39 & 12.85 & 41.37 & 24.67 & 13.58 & 10.59 & 4.40 & 14.50 & 00 & 00 & 01 \\
 & IForest & 8.59 & 0.62 & 23.57 & 56.50 & 11.56 & 7.71 & 70.99 & 46.62 & 2.88 & 2.06 & 10.47 & 11.29 & 15.85 & 7.55 & 8.88 & 15.49 & 00 & 00 & 00 \\
 & Sub-PCA & 23.02 & 0.60 & \smedal{} 46.08 & \textbf{\gmedal{} 88.91} & 10.49 & 3.72 & 80.86 & 54.16 & 12.92 & 16.41 & 21.57 & 31.43 & 18.52 & 26.42 & 37.50 & 43.02 & 01 & 01 & 00 \\
 & DCdetector & 5.83 & 0.59 & 16.60 & 14.03 & 12.32 & 9.37 & 74.16 & 46.66 & 1.53 & 3.23 & 10.17 & 7.01 & 14.49 & 4.21 & 4.66 & 15.04 & 00 & 00 & 00 \\
 & TFMAE & 5.32 & 0.64 & 15.68 & 17.81 & 11.90 & 9.55 & 73.54 & 48.79 & 2.57 & 5.36 & 25.93 & 8.25 & 14.22 & 5.76 & 4.77 & 15.38 & 00 & 00 & 00 \\
        \hline
    \end{tabular}
    }
    \vspace{-10pt}
\end{table*}

\clearpage
\section{Closest-method comparison on multivariate families}
\label{app:closest-mv}

The frequency-domain and feature-based detectors compared in Section~\ref{sec:closest} are univariate by construction.
For completeness we also apply them to the multivariate families by scoring each channel independently and averaging the per-channel scores (catch22 features are likewise computed per channel).
Table~\ref{tab:closest-mv} reports this extension: the discovered multivariate detector remains ahead on all four metrics, most clearly on VUS-PR.
Because these are univariate methods applied per channel rather than native multivariate detectors, we treat the comparison as indicative.

\begin{table}[htbp]
  \centering
  \caption{Comparison with the closest published methods on the TSB-AD multivariate families, applying each univariate baseline per channel and averaging (Time-RCD protocol; dataset-mean, higher is better $\uparrow$).
  Best per column in \textbf{bold}.}
  \label{tab:closest-mv}
  \small
  \setlength{\tabcolsep}{5pt}
  \begin{tabular}{lcccc}
    \toprule
    Method & Aff-F\,$\uparrow$ & F1-T\,$\uparrow$ & Std-F1\,$\uparrow$ & VUS-PR\,$\uparrow$ \\
    \midrule
    \textbf{Opus 4.8 Loop (ours)} & \textbf{85.79} & \textbf{49.93} & \textbf{46.62} & \textbf{35.16} \\
Spectral Residual~\citep{ren2019time} & 84.17 & 36.32 & 36.12 & 27.33 \\
C22MP~\citep{tafazoli2023c22mp} & 75.46 & 33.18 & 29.09 & 18.27 \\
FFT detector~\citep{rasheed2009fft} & 77.24 & 29.83 & 28.34 & 17.05 \\
catch22 + Mahalanobis~\citep{lubba2019catch22} & 75.73 & 31.81 & 27.15 & 15.82
 \\
    \bottomrule
  \end{tabular}
\end{table}

\end{document}